\documentclass[conference]{IEEEtran}
\IEEEoverridecommandlockouts

\usepackage{cite}
\usepackage{amsmath}
\usepackage{amssymb}
\usepackage{amsfonts}
\usepackage{graphicx}
\usepackage{textcomp}
\usepackage{xcolor}

\usepackage{algorithm}
\usepackage{algpseudocode}
\usepackage{mwe}
\usepackage{booktabs}
\usepackage{makecell}
\usepackage{subcaption}
\usepackage{array}
\usepackage{arydshln}

\def\BibTeX{{\rm B\kern-.05em{\sc i\kern-.025em b}\kern-.08em
    T\kern-.1667em\lower.7ex\hbox{E}\kern-.125emX}}

\DeclareMathOperator*{\argmin}{arg\,min}
\DeclareMathOperator*{\argmax}{arg\,max}

\newcommand{\model}{PerbALGraph\xspace}

\begin{document}

\title{Perturbation-based Graph Active Learning for Weakly-Supervised Belief Representation Learning}

\author{
\IEEEauthorblockN{
Dachun Sun,
Ruijie Wang,
Jinning Li,
Ruipeng Han,
Xinyi Liu,
You Lyu,
and Tarek Abdelzaher
}
\IEEEauthorblockA{
Department of Computer Science\\
University of Illinois at Urbana-Champaign, Urbana IL 61801, USA\\
\{dsun18, ruijiew2, jinning4, ruipeng2, liu323, youlyu2, zaher\}@illinois.edu
}
}

\maketitle

\begin{abstract}
This paper addresses the problem of optimizing the allocation of labeling resources for semi-supervised belief representation learning in social networks.
The objective is to strategically identify valuable messages on social media graphs that are worth labeling within a constrained budget, ultimately maximizing the task's performance.
Despite the progress in unsupervised or semi-supervised methods in advancing belief and ideology representation learning on social networks and the remarkable efficacy of graph learning techniques, the availability of high-quality curated labeled social data can greatly benefit and further improve performances.
Consequently, allocating labeling efforts is a critical research problem in scenarios where labeling resources are limited. 
This paper proposes a graph data augmentation-inspired perturbation-based active learning strategy (\model) that progressively selects messages for labeling according to an automatic estimator, obviating human guidance. 
This estimator is based on the principle that messages in the network that exhibit heightened sensitivity to structural features of the observational data indicate landmark quality that significantly influences semi-supervision processes.
We design the estimator to be the prediction variance under a set of designed graph perturbations, which is model-agnostic and application-independent. 
Extensive experiment results demonstrate the effectiveness of the proposed strategy for belief representation learning tasks.
\end{abstract}


\section{Introduction}

As social networks become ever more central to human society, individuals' online behaviors increasingly reveal their beliefs, such as stances, viewpoints, and ideological preferences. To capture these insights, belief representation learning is proposed to uncover latent structures in a low-dimensional space that represent these beliefs based on observable online behaviors on social media~\cite{xiao2020timme,yang2022hierarchicaloverlappingbeliefestimation,li2022unsupervised,10.3389/fdata.2021.729881,li2024large}.

State-of-the-art methods deploy Graph Neural Networks (GNNs) to derive unsupervised belief representations from online behavioral graphs (e.g., who shares or posts what)~\cite{li2022unsupervised}, enabling advanced applications like classifying users and posts by their beliefs~\cite{yang2022hierarchicaloverlappingbeliefestimation}, forecasting stances on various topics~\cite{dong2017weakly}, and identifying polarization within belief spaces~\cite{al2017unveiling} . By concentrating exclusively on graph topology rather than post content or downstream task labels, these methods get rid of expensive labeling efforts and exhibit remarkable generalization across textual~\cite{li2022unsupervised,10.3389/fdata.2021.729881,li2024large}, visual~\cite{liu2023unsupervisedimageclassificationideological}, and multi-modal belief representation learning~\cite{10429919}.

While unsupervised GNN-based methods offer several advantages, they encounter limitations in specific real-world scenarios that demand weak supervision. In social networks, for instance, the inherent power law distribution creates challenges in accurately inferring beliefs for nodes within isolated or sparse components, where the structural inductive biases learned through unsupervised methods prove ineffective. Additionally, when a pre-defined belief axis exists, supervision becomes essential for aligning the latent space with this axis. For example, in discussions surrounding geopolitical events, supervision is necessary to categorize users and posts along a predefined axis such as Conservatism versus Liberalism, ensuring that viewpoints are correctly aligned with their respective ideological stances. 

Despite the abundance of unlabeled data in social networks and the progress in GNNs, obtaining labeled data for belief representation learning remains challenging due to limited budgets or resources~\cite{li2024large}. Therefore, allocating labeling efforts strategically to representative nodes within a constrained budget is critical for improving performances in various applications. Active learning~\cite{settles2009active} is a well-established research area in machine learning, which aims to select the most informative instances for labeling from a large pool of unlabeled data. However, many existing methods extract limited information during the training process and graph structure itself~\cite{cai2017active,chen2019activehneactiveheterogeneousnetwork,anrmab}, which may not be effective for diverse applications.

We introduce a novel approach called \model to extract deeper insights from both the learning model and the graph structure by leveraging input graph variations through targeted perturbations. These perturbations, specifically tailored for graph-structured data, include techniques such as edge dropping, noisy edge addition, and path dropping. By analyzing the model’s inferences on these perturbed graphs, we gain a more comprehensive understanding of both the model’s behavior and the underlying graph structure, enabling more robust and informed decision-making in graph-based learning tasks.

To evaluate the effectiveness of our proposed active learning strategy, we conduct experiments on various types of GNN-based encoders and node classification tasks, including ideology identification in social media data. We also empirically examine the selected nodes in the social media data to reveal more insights. Our experiment results show that the proposed approach outperforms existing methods in terms of classification accuracy and labeling efficiency, indicating that it is a promising direction for optimizing the labeling budget allocation problem in semi-supervised graph learning.

To sum up, our contributions are three-fold:
\begin{itemize}
    \item To the best of our knowledge, we are the first to study graph active learning framework to effectively address weakly-supervised belief representation learning problem.
    \item We propose a graph active learning algorithm that quantifies uncertainty through performance variance induced by graph perturbations. This approach efficiently selects nodes from social networks for labeling, facilitating more effective belief representation learning.
    \item We conduct extensive experiments and empirical observations on six social media datasets. The results demonstrate the superority of the proposed \model framework.
\end{itemize}

The paper is structured as follows. Section~\ref{sec:definition} states problem definition. Section~\ref{sec:method} introduces our proposed method, followed by the experiment results and analysis in Section~\ref{sec:exp}. 
We review literature related to graph learning, active learning, and social media data mining in Section~\ref{sec:related}, and conclude with a summary in Section~\ref{sec:conclusion}.

\section{Problem Definition}\label{sec:definition}
We consider a graph $\mathcal{G}=(\mathcal{V}, \mathcal{E})$ with $\left|\mathcal{V}\right|=n$, number of node classes $c$, adjacency matrix $\mathbf{A}\in\{0,1\}^{n\times n}$ ($\mathbf{A}\in\mathbb{R}^{n\times n}$ if the graph is weighted), and node features $\mathbf{X}=\left[\mathbf{x}^\top_1, \mathbf{x}^\top_2, ..., \mathbf{x}^\top_n\right]$, where $x_i\in \mathbb{R}^d$ and $d$ is the feature dimension. Node set $\mathcal{V}$ is split into  training set $\mathcal{V}_{\text{train}}$, validation set $\mathcal{V}_{\text{val}}$, and testing set $\mathcal{V}_{\text{test}}$. Given a budget $B$, a labeling oracle that can assign a node $v_i$ a label $y_i\in\{0,1\}^c$, and a set of initially labeled nodes $\mathcal{V}_L\subseteq\mathcal{V}_{\text{train}}$ ($n_L=\left|\mathcal{V}_L\right|$), the objective is to optimize the performance of the learning model by designing an active learning selection strategy to select $(B-n_L)$ nodes from the unlabeled node set $\mathcal{V}_U$ for oracle to label and append to $\mathcal{V}_L$:
\begin{equation}
\argmin_{\mathcal{V}_L:\left|\mathcal{V}_L\right|=B}\mathbb{E}_{v_i\in\mathcal{V}_{\text{test}}}\left[\mathcal{L}\left(y_i, P(\Hat{y_i}|f_\phi(G;\mathcal{V}_L))\right)\right]
\end{equation}
where $P(\Hat{y_i}|f_\phi)$ is the predicted label distribution of node $v_i$ by the learning model $f_\phi$ (e.g., GCN).

In the context of social data mining and belief representation learning, assertions are collections of messages claiming the same idea. $\mathcal{V}$ consists of assertions and users, and $\mathcal{G}$ becomes a bipartite graph. $\mathbf{X}$ can be derived from assertion texts, from the label, or be the identity matrix. Furthermore, the chosen nodes will be examined empirically and could provide insight into the specific social media data.

We follow the setting of a pool-based active learning on the previously defined graph. Initially, only the labeled set $\mathcal{V}_L, \left\{y_i\right\}_{i\in\mathcal{V}_L}$ are revealed to the learning model. Our method then identifies a query node set $\mathcal{V}_Q$ from $\mathcal{V}_U$ and queries the labeling oracle for labels. These node sets are correspondingly updated ($\mathcal{V}_L\leftarrow\mathcal{V}_L\cup\mathcal{V}_Q$, $\mathcal{V}_U\leftarrow\mathcal{V}_U$\textbackslash$\mathcal{V}_Q$) until the labeling budget is depleted. The size $b=\left|\mathcal{V}_Q\right|$, as a hyperparameter, decides whether we conduct sequential or batch active learning.

\section{Methodology}\label{sec:method}

\begin{figure}
    \centering
    \includegraphics[width=0.9\linewidth]{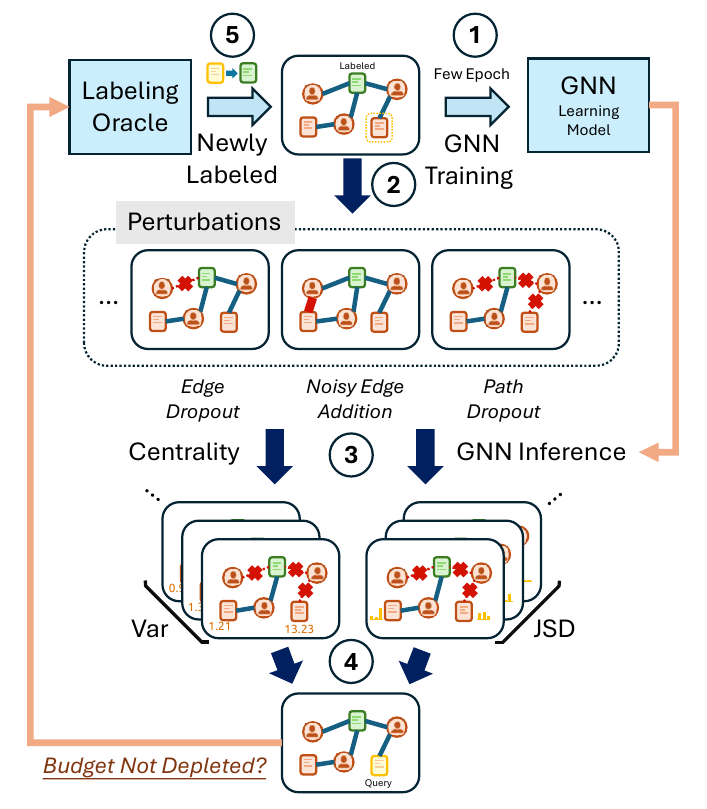}
    \caption{Overview of \model framwork.}
    \label{fig:enter-label}
\end{figure}

\subsection{Perturbation-based Graph Active-Learning}

Previous methods focus on how to extract the informativeness or representativeness of a node on the original graph $\mathcal{G}$ with the latest learning model. Predicted label distribution tends to be the most meaningful information. Inspired by works on graph data augmentation and graph contrastive learning~\cite{zhao2021data}, we propose to extract more from the learning model and the graph structure itself using variations of the input graph by applying perturbations designed for graph-structured data, including edge dropping, noisy edge addition, and path dropping. Inferences by the learning model on perturbed graphs can offer more insights about the learning model itself and the graph structure.

After performing these operations $a_e$, $a_m$, and $a_p$ times respectively, we obtain a set of perturbed graphs $\{\Tilde{\mathcal{G}}_i\}_{i\in [1,\mathbf{a}]}$, where $\mathbf{a}=\sum a_*$. Our method calculates several metrics for each unlabeled node and identifies a query node set.

\subsubsection{Instability} To elicit more information from the learning model and the graph structure, we compute the predicted label distributions for unlabeled nodes in each perturbed graph, 
$\left\{\mathbf{P}_i\right\}_{v_i\in\mathcal{V}_U}$, where each $\mathbf{P}_i=\left[P\left(\Hat{y_{i}}|f_\phi(\Tilde{\mathcal{G}}_j;\mathcal{V}_L)\right)\right]_{j\in[1,\mathbf{a}]}$ consists of predicted label distributions of $v_i$ evaluated by the learning model on each perturbed graph. To be more specific, each $\mathbf{P}_{i,j}$ is a label distribution of node $v_i$ inferred by the learning model on graph $\Tilde{\mathcal{G}}_j$. We define the \textbf{instability} of node $v_i$,
\begin{equation}\label{eq:instability}
    \phi_\text{instability}(v_i)=\text{JSD}(\mathbf{P}_i) \\
\end{equation}
where JSD is the generalized Jensen-Shannon divergence,
\begin{equation}
    \text{JSD}(\mathbf{P}_*)=H\left(\frac{1}{n}\sum_{i=1}^n \mathbf{P}_{*,i}\right)-\frac{1}{n}\sum_{i=1}^n H(\mathbf{P}_{*,i})
\end{equation}
and $H(P)$ is the Shannon entropy. The key observation is when the node has a high instability, meaning that the perturbed graph structure greatly changed the predicted label distribution, the node is then highly suitable to be chosen as an anchor.

\subsubsection{Sensitivity} In social network analysis, many metrics to measure the centrality of a node have been proposed. We adopt betweenness centrality~\cite{freeman1977set} and an eigenvector-based PageRank centrality~\cite{pagerankalg}, and select the appropriate one to use based on the data. PageRank centrality of $v_i$ is calculated as
\begin{equation}
    \mu_\text{PageRank}(v_i;\mathcal{G})=\rho\sum_{j=1}^n \mathbf{A}_{ij}\frac{ \mu_\text{PageRank}(v_j;\mathcal{G})}{\sum_{k=1}^n \mathbf{A}_{jk}}+\frac{1-\rho}{n}
\end{equation}
and the betweenness centrality of $v_i$ is calculated as
\begin{equation}
    \mu_\text{betweenness}(v_i;\mathcal{G})=\sum_{\substack{s,v,t\in\mathcal{V}_\mathcal{G} \\ s\neq v\neq t}}\frac{\sigma_{st}(v)}{\sigma_{st}}
\end{equation}
where $\sigma_{st}$ is the total number of shortest paths from $s$ to $t$, and $\sigma_{st}(v)$ is the number of those paths that pass through $v$.

To gain more information about the graph structure, we also compute the centrality metrics for each perturbed graph. We calculate the centrality metrics for each unlabeled node on all perturbed graphs, giving $\left\{\boldsymbol{\mu}^\text{(centrality)}_i\right\}_{v_i\in\mathcal{V}_U}$, 
where each $\boldsymbol{\mu}^\text{(centrality)}_i=\left[\mu_\text{centrality}(v_i;\mathcal{G}_j)\right]_{j\in[1,\mathbf{a}]}$
We define the \textbf{sensitivity} of a node $v_i$ to centrality metric $m$ as
\begin{equation}\label{eq:sensitivity}
    \phi_\text{sensitivity}(v_i)=\text{Var}\left[\boldsymbol{\mu}^\text{(m)}_i\right]
\end{equation}

\subsubsection{Score Combination} Since different metrics above are of incomparable scale, we first convert them into percentiles as in \cite{scorenormal}. Define $\text{perc}_m(v, \mathcal{V}_U)$ as the percentile of nodes in $\mathcal{V}_U$ that have smaller values than node $v$ according to metric $m$. Then, the objective function of our proposed method to select a node for labeling is:
\begin{align}
    v^*=\argmax_{v\in\mathcal{V}_U}\biggr[ &\gamma_t\cdot\text{perc}_\text{instability}(v, \mathcal{V}_U)\nonumber \\ 
    & + (1-\gamma_t)\cdot \text{perc}_\text{sensitivity}(v, \mathcal{V}_U)\biggr]
\end{align}
or in the case of selecting a batch of nodes:
\begin{align}
    V^*=\argmax_{\substack{V\subset\mathcal{V}_U, \\ \left|V\right|=b}}\sum_{v\in V}\biggr[ &\gamma_t\cdot\text{perc}_\text{instability}(v, \mathcal{V}_U)\nonumber \\ 
    & + (1-\gamma_t)\cdot \text{perc}_\text{sensitivity}(v, \mathcal{V}_U)\biggr]\label{eq:batch_al}
\end{align}

We follow the settings in \cite{cai2017active} and \cite{zhang2017active} where $\gamma_t$ is a random variable between 0 and 1 with a temporal dependence on AL iteration, and it is responsible for shifting the attention from the structure-based sensitivity score to learning-and-uncertainty-based instability score. We sample $\gamma_t \sim \text{Beta}(1, \beta_t)$. During the AL iteration, we decrease $\beta_t$ linearly so that the expectation of $\gamma_t$ increases, aligning with the need for increasing attention on the instability score.

\subsubsection{Graph Perturbation}
As a data augmentation method mainly to facilitate contrastive learning on graphs \cite{graphcontrastive}, we apply the following augmentations to elicit the noteworthy node candidates from the learning model.
\begin{itemize}
    \item Edge dropping: Randomly drops edges from the adjacency matrix with probability $p$ using samples from a Bernoulli distribution.
    \item Add random edge: Randomly add edges to the adjacency matrix.
    \item Path dropping: Drops edges from the adjacency matrix based on random walks.
\end{itemize}

We do not employ node dropping because the underlying prior for it is that missing a vertex does not alter semantics~\cite{graphcontrastive}.

\begin{algorithm}[t]
\caption{Perturbation-based Graph Active-Learning}\label{alg:main}
\begin{algorithmic}
\Require Budget $B$, graph $\mathcal{G}$, initial labeled node set $\mathcal{V}_L$, candidate node set $\mathcal{V}_U$, query node set size $b$, perturbation configuration $(a_e, a_m, a_p)$
\Ensure $\mathcal{V}_\text{selected}$
\State $\mathcal{V}_\text{selected} \gets \mathcal{V}_L$
\State $t \gets 1$
\While {budget not depleted}
  \State Train the learning model $f_\phi(G;\mathcal{V}_L))$
  \State Form the perturbed graphs $\{\Tilde{\mathcal{G}}_i\}_{i\in [1,a_e+a_m+a_p]}$
  \State Run learning model inference to obtain $\left\{\mathbf{P}_i\right\}_{v_i\in\mathcal{V}_U}$
  \State Obtain $\left\{\boldsymbol{\mu}^\text{(centrality)}_i\right\}_{v_i\in\mathcal{V}_U}$ for each perturbed graph.
  \State Calculate scores with Eq~\ref{eq:instability} and Eq~\ref{eq:sensitivity}.
  \State Sample $\gamma_t \sim \text{Beta}(1, \beta_t)$, and calculate a combined score for each node in $\mathcal{V}_U$.
  \State Calculate $V^*$ in Eq~\ref{eq:batch_al}
  \State $\mathcal{V}_\text{selected} \gets \mathcal{V}_\text{selected}\bigcup V^*$
  \State $B \gets B - b$
  \State $t \gets t + 1$
  \State Decrease $\beta_t$
\EndWhile
\State \Return $\mathcal{V}_\text{selected}$
\end{algorithmic}
\end{algorithm}

\section{Experiments}\label{sec:exp}

In this section, we evaluate two variants of the proposed perturbation-based AL methods, alongside nine baseline models, using the belief representation learning task across six real-world X (formerly Twitter) datasets. The evaluation is conducted under five fixed labeling budgets (5, 10, 15, 20, and 30), allowing us to investigate the impact of budget size on downstream tasks.

\begin{table}[t]
\caption{Dataset Statistics. The average degree represents the overall sparsity. Dataset names are abbreviated and ``claims'' are synonymous to ``assertions''.}\label{table:datasets}
\centering
\renewcommand\arraystretch{1.1}
\begin{tabular}{cccccc}
\toprule
Dataset & \#Posts & \#Claims & \#Users & \#Edges & Avg. Deg. \\
\hline
Eurovision & 3371 & 537 & 992 & 3081 & 4.03 \\
\hline
Russia/Ukraine & 17153 & 1582 & 1577 & 5519 & 3.49 \\
\hline
EDCA & 5219 & 467 & 912 & 4916 & 7.13 \\
\hline
Sovereignty & 101340 & 4386 & 6807 & 98302 & 17.56 \\
\hline
Energy Issues & 5950 & 189 & 3047 & 5939 & 3.67 \\
\hline
Separatism & 5474 & 750 & 957 & 5137 & 6.02 \\
\bottomrule
\end{tabular}
\end{table}

We focus on the performance of our proposed method with an emphasis on scenarios with a constrained labeling budget of 20 nodes. This includes a detailed performance table and an examination of the queried nodes during the active learning process for two datasets. Meanwhile, we also perform a comparative analysis of our method against other active learning methods across a range of labeling budgets. 

As the perturbation-based method involves creating multiple copies of the input graph, we also evaluate its computational efficiency and scalability. Finally, we conduct an ablation study to assess the effect of varying the number of perturbed graphs on the overall performance.

\begin{table*}[htb]
\caption{Evaluation results for belief representation learning using GCN and SGVGAE trained in a semi-supervised fashion on AL-queried nodes \textbf{with a budget of 20 queries}. The number next to the dataset name represents the percentage of the budget relative to the number of candidates in the dataset.}\label{table:eval}
\centering
\renewcommand\arraystretch{1.15}
\begin{tabular}{l|cccc|cccc}
\hline
\multicolumn{1}{r|}{Dataset} & \multicolumn{4}{c|}{Eurovision [3.7\%]} & \multicolumn{4}{c}{EDCA (Philippines) [4.3\%]}\\
\cline{2-9}
\multicolumn{1}{r|}{Belief Embedding} & \multicolumn{2}{c|}{GCN} & \multicolumn{2}{c|}{SGVGAE} & \multicolumn{2}{c|}{GCN} & \multicolumn{2}{c}{SGVGAE}\\
\cline{2-9}
AL Method & Acc. (\%) & \multicolumn{1}{c|}{Macro F1 (\%)} & Acc. (\%) & Macro F1 (\%) & Acc. (\%) & \multicolumn{1}{c|}{Macro F1 (\%)} & Acc. (\%) & Macro F1 (\%) \\
\hline
Random & 77.25$^{\scriptscriptstyle{\pm}\text{7.14}}$ & 74.23$^{\scriptscriptstyle{\pm}\text{8.05}}$ & 76.36$^{\scriptscriptstyle{\pm}\text{5.22}}$ & 75.29$^{\scriptscriptstyle{\pm}\text{4.87}}$ & 83.75$^{\scriptscriptstyle{\pm}\text{2.03}}$ & 77.17$^{\scriptscriptstyle{\pm}\text{3.45}}$ & 65.91$^{\scriptscriptstyle{\pm}\text{14.01}}$ & 63.12$^{\scriptscriptstyle{\pm}\text{14.93}}$ \\
Centrality (Degree) & \underline{83.47}$^{\scriptscriptstyle{\pm}\text{1.77}}$ & \underline{80.73}$^{\scriptscriptstyle{\pm}\text{1.76}}$ & 74.00$^{\scriptscriptstyle{\pm}\text{4.82}}$ & 72.88$^{\scriptscriptstyle{\pm}\text{4.58}}$ & 84.24$^{\scriptscriptstyle{\pm}\text{0.59}}$ & 76.86$^{\scriptscriptstyle{\pm}\text{1.37}}$ & 65.53$^{\scriptscriptstyle{\pm}\text{13.89}}$ & 63.88$^{\scriptscriptstyle{\pm}\text{13.72}}$ \\
Centrality (PageRank) & 82.40$^{\scriptscriptstyle{\pm}\text{4.79}}$ & 80.16$^{\scriptscriptstyle{\pm}\text{4.73}}$ & 75.09$^{\scriptscriptstyle{\pm}\text{5.32}}$ & 73.92$^{\scriptscriptstyle{\pm}\text{5.12}}$ & 80.36$^{\scriptscriptstyle{\pm}\text{1.44}}$ & 74.88$^{\scriptscriptstyle{\pm}\text{1.56}}$ & \underline{71.21}$^{\scriptscriptstyle{\pm}\text{13.23}}$ & \underline{69.67}$^{\scriptscriptstyle{\pm}\text{13.14}}$ \\
Centrality (Betweenness) & 82.13$^{\scriptscriptstyle{\pm}\text{3.58}}$ & 78.00$^{\scriptscriptstyle{\pm}\text{3.60}}$ & 76.00$^{\scriptscriptstyle{\pm}\text{5.91}}$ & 74.81$^{\scriptscriptstyle{\pm}\text{5.78}}$ & 83.07$^{\scriptscriptstyle{\pm}\text{3.08}}$ & 74.18$^{\scriptscriptstyle{\pm}\text{7.86}}$ & 67.80$^{\scriptscriptstyle{\pm}\text{11.80}}$ & 65.70$^{\scriptscriptstyle{\pm}\text{11.82}}$ \\
Entropy & 78.39$^{\scriptscriptstyle{\pm}\text{5.28}}$ & 73.36$^{\scriptscriptstyle{\pm}\text{6.09}}$ & 76.48$^{\scriptscriptstyle{\pm}\text{5.10}}$ & 75.31$^{\scriptscriptstyle{\pm}\text{4.81}}$ & 83.57$^{\scriptscriptstyle{\pm}\text{3.00}}$ & 78.44$^{\scriptscriptstyle{\pm}\text{3.38}}$ & 61.36$^{\scriptscriptstyle{\pm}\text{12.96}}$ & 58.20$^{\scriptscriptstyle{\pm}\text{13.37}}$ \\
AGE & 79.07$^{\scriptscriptstyle{\pm}\text{4.83}}$ & 76.31$^{\scriptscriptstyle{\pm}\text{4.77}}$ & 76.00$^{\scriptscriptstyle{\pm}\text{5.77}}$ & 74.90$^{\scriptscriptstyle{\pm}\text{5.48}}$ & \underline{84.99}$^{\scriptscriptstyle{\pm}\text{2.86}}$ & \textbf{80.81}$^{\scriptscriptstyle{\pm}\text{2.67}}$ & 61.36$^{\scriptscriptstyle{\pm}\text{12.96}}$ & 58.20$^{\scriptscriptstyle{\pm}\text{13.37}}$ \\
ANRMAB & 79.28$^{\scriptscriptstyle{\pm}\text{6.70}}$ & 76.06$^{\scriptscriptstyle{\pm}\text{7.08}}$ & 77.00$^{\scriptscriptstyle{\pm}\text{5.55}}$ & 75.83$^{\scriptscriptstyle{\pm}\text{5.28}}$ & 79.32$^{\scriptscriptstyle{\pm}\text{4.61}}$ & 72.95$^{\scriptscriptstyle{\pm}\text{4.52}}$ & 61.36$^{\scriptscriptstyle{\pm}\text{12.96}}$ & 58.20$^{\scriptscriptstyle{\pm}\text{13.37}}$ \\
GPA & 75.70$^{\scriptscriptstyle{\pm}\text{6.62}}$ & 73.74$^{\scriptscriptstyle{\pm}\text{5.94}}$ & 77.26$^{\scriptscriptstyle{\pm}\text{4.86}}$ & 76.08$^{\scriptscriptstyle{\pm}\text{4.58}}$ & 80.67$^{\scriptscriptstyle{\pm}\text{2.24}}$ & 73.30$^{\scriptscriptstyle{\pm}\text{2.08}}$ & 63.18$^{\scriptscriptstyle{\pm}\text{12.80}}$ & 61.50$^{\scriptscriptstyle{\pm}\text{12.62}}$ \\
RIM & 70.00$^{\scriptscriptstyle{\pm}\text{0.00}}$ & 41.18$^{\scriptscriptstyle{\pm}\text{0.00}}$ & 77.00$^{\scriptscriptstyle{\pm}\text{4.63}}$ & 75.95$^{\scriptscriptstyle{\pm}\text{4.32}}$ & 77.12$^{\scriptscriptstyle{\pm}\text{4.25}}$ & 68.56$^{\scriptscriptstyle{\pm}\text{3.85}}$ & 65.53$^{\scriptscriptstyle{\pm}\text{13.89}}$ & 63.88$^{\scriptscriptstyle{\pm}\text{13.72}}$ \\ \hdashline
Ours (PageRank) & \textbf{85.14}$^{\scriptscriptstyle{\pm}\text{5.64}}$ & \textbf{81.71}$^{\scriptscriptstyle{\pm}\text{6.34}}$ & \textbf{81.33}$^{\scriptscriptstyle{\pm}\text{6.43}}$ & \textbf{79.98}$^{\scriptscriptstyle{\pm}\text{6.40}}$ & \textbf{86.04}$^{\scriptscriptstyle{\pm}\text{2.76}}$ & \underline{79.84}$^{\scriptscriptstyle{\pm}\text{4.66}}$ & \textbf{73.86}$^{\scriptscriptstyle{\pm}\text{11.25}}$ & \textbf{71.21}$^{\scriptscriptstyle{\pm}\text{12.11}}$ \\
Ours (Betweenness) & 82.86$^{\scriptscriptstyle{\pm}\text{4.14}}$ & 78.57$^{\scriptscriptstyle{\pm}\text{5.94}}$ & \underline{79.50}$^{\scriptscriptstyle{\pm}\text{6.61}}$ & \underline{78.38}$^{\scriptscriptstyle{\pm}\text{6.30}}$ & 79.55$^{\scriptscriptstyle{\pm}\text{4.15}}$ & 72.36$^{\scriptscriptstyle{\pm}\text{6.01}}$ & \textbf{73.86}$^{\scriptscriptstyle{\pm}\text{11.25}}$ & \textbf{71.21}$^{\scriptscriptstyle{\pm}\text{12.11}}$ \\
\hline
\hline
\multicolumn{1}{r|}{Dataset} & \multicolumn{4}{c|}{Russia/Ukraine Conflict (Visual) [1.3\%]} & \multicolumn{4}{c}{Territorial Sovereignty (Philippines) [0.5\%]}\\
\cline{2-9}
\multicolumn{1}{r|}{Belief Embedding} & \multicolumn{2}{c|}{GCN} & \multicolumn{2}{c|}{SGVGAE} & \multicolumn{2}{c|}{GCN} & \multicolumn{2}{c}{SGVGAE}\\
\cline{2-9}
AL Method & Acc. (\%) & \multicolumn{1}{c|}{Macro F1 (\%)} & Acc. (\%) & Macro F1 (\%) & Acc. (\%) & \multicolumn{1}{c|}{Macro F1 (\%)} & Acc. (\%) & Macro F1 (\%) \\
\hline
Random & 64.06$^{\scriptscriptstyle{\pm}\text{2.76}}$ & 53.67$^{\scriptscriptstyle{\pm}\text{9.33}}$ & 80.71$^{\scriptscriptstyle{\pm}\text{2.79}}$ & 80.50$^{\scriptscriptstyle{\pm}\text{2.87}}$ & \underline{83.11}$^{\scriptscriptstyle{\pm}\text{0.00}}$ & 46.69$^{\scriptscriptstyle{\pm}\text{0.00}}$ & 69.41$^{\scriptscriptstyle{\pm}\text{11.49}}$ & 57.11$^{\scriptscriptstyle{\pm}\text{8.23}}$ \\
Centrality (Degree) & 67.27$^{\scriptscriptstyle{\pm}\text{0.95}}$ & 55.80$^{\scriptscriptstyle{\pm}\text{1.87}}$ & 81.25$^{\scriptscriptstyle{\pm}\text{2.89}}$ & 81.03$^{\scriptscriptstyle{\pm}\text{2.98}}$ & 82.88$^{\scriptscriptstyle{\pm}\text{0.00}}$ & 45.32$^{\scriptscriptstyle{\pm}\text{0.00}}$ & 66.77$^{\scriptscriptstyle{\pm}\text{12.42}}$ & 56.15$^{\scriptscriptstyle{\pm}\text{7.82}}$ \\
Centrality (PageRank) & 59.89$^{\scriptscriptstyle{\pm}\text{10.40}}$ & 57.31$^{\scriptscriptstyle{\pm}\text{12.17}}$ & 80.24$^{\scriptscriptstyle{\pm}\text{2.80}}$ & 80.02$^{\scriptscriptstyle{\pm}\text{2.92}}$ & 82.88$^{\scriptscriptstyle{\pm}\text{0.00}}$ & 45.32$^{\scriptscriptstyle{\pm}\text{0.00}}$ & 69.00$^{\scriptscriptstyle{\pm}\text{11.70}}$ & 55.10$^{\scriptscriptstyle{\pm}\text{6.87}}$ \\
Centrality (Betweenness) & \underline{67.58}$^{\scriptscriptstyle{\pm}\text{9.76}}$ & \underline{65.27}$^{\scriptscriptstyle{\pm}\text{8.70}}$ & 80.68$^{\scriptscriptstyle{\pm}\text{2.99}}$ & 80.51$^{\scriptscriptstyle{\pm}\text{3.04}}$ & 82.88$^{\scriptscriptstyle{\pm}\text{0.00}}$ & 45.32$^{\scriptscriptstyle{\pm}\text{0.00}}$ & 72.98$^{\scriptscriptstyle{\pm}\text{9.79}}$ & 58.51$^{\scriptscriptstyle{\pm}\text{8.56}}$ \\
Entropy & 65.83$^{\scriptscriptstyle{\pm}\text{1.89}}$ & 55.58$^{\scriptscriptstyle{\pm}\text{5.62}}$ & 80.64$^{\scriptscriptstyle{\pm}\text{2.30}}$ & 80.37$^{\scriptscriptstyle{\pm}\text{2.44}}$ & \textbf{83.21}$^{\scriptscriptstyle{\pm}\text{0.14}}$ & 47.29$^{\scriptscriptstyle{\pm}\text{0.81}}$ & 71.53$^{\scriptscriptstyle{\pm}\text{10.51}}$ & 58.66$^{\scriptscriptstyle{\pm}\text{8.00}}$ \\
AGE & 60.24$^{\scriptscriptstyle{\pm}\text{0.11}}$ & 40.14$^{\scriptscriptstyle{\pm}\text{0.28}}$ & 79.87$^{\scriptscriptstyle{\pm}\text{4.20}}$ & 79.62$^{\scriptscriptstyle{\pm}\text{4.30}}$ & 77.63$^{\scriptscriptstyle{\pm}\text{7.50}}$ & 46.58$^{\scriptscriptstyle{\pm}\text{1.90}}$ & 69.60$^{\scriptscriptstyle{\pm}\text{10.64}}$ & 58.08$^{\scriptscriptstyle{\pm}\text{7.88}}$ \\
ANRMAB & 65.09$^{\scriptscriptstyle{\pm}\text{3.31}}$ & 58.33$^{\scriptscriptstyle{\pm}\text{5.19}}$ & 80.65$^{\scriptscriptstyle{\pm}\text{3.91}}$ & 80.39$^{\scriptscriptstyle{\pm}\text{4.03}}$ & 82.88$^{\scriptscriptstyle{\pm}\text{0.00}}$ & 45.32$^{\scriptscriptstyle{\pm}\text{0.00}}$ & 69.07$^{\scriptscriptstyle{\pm}\text{11.54}}$ & 56.48$^{\scriptscriptstyle{\pm}\text{8.10}}$ \\
GPA & 61.10$^{\scriptscriptstyle{\pm}\text{7.85}}$ & 56.50$^{\scriptscriptstyle{\pm}\text{9.74}}$ & 81.20$^{\scriptscriptstyle{\pm}\text{3.84}}$ & 80.96$^{\scriptscriptstyle{\pm}\text{3.96}}$ & 67.83$^{\scriptscriptstyle{\pm}\text{11.18}}$ & \underline{51.67}$^{\scriptscriptstyle{\pm}\text{3.58}}$ & 67.67$^{\scriptscriptstyle{\pm}\text{10.92}}$ & 55.89$^{\scriptscriptstyle{\pm}\text{7.14}}$ \\
RIM & 43.18$^{\scriptscriptstyle{\pm}\text{0.00}}$ & 34.52$^{\scriptscriptstyle{\pm}\text{0.00}}$ & 80.28$^{\scriptscriptstyle{\pm}\text{3.35}}$ & 80.04$^{\scriptscriptstyle{\pm}\text{3.43}}$ & 82.88$^{\scriptscriptstyle{\pm}\text{0.00}}$ & 45.32$^{\scriptscriptstyle{\pm}\text{0.00}}$ & 69.71$^{\scriptscriptstyle{\pm}\text{11.16}}$ & 56.95$^{\scriptscriptstyle{\pm}\text{7.64}}$ \\ \hdashline
Ours (PageRank) & 67.40$^{\scriptscriptstyle{\pm}\text{4.87}}$ & 64.14$^{\scriptscriptstyle{\pm}\text{5.42}}$ & \textbf{82.58}$^{\scriptscriptstyle{\pm}\text{1.71}}$ & \textbf{82.52}$^{\scriptscriptstyle{\pm}\text{1.73}}$ & 68.95$^{\scriptscriptstyle{\pm}\text{12.72}}$ & \textbf{53.75}$^{\scriptscriptstyle{\pm}\text{6.18}}$ & \textbf{78.15}$^{\scriptscriptstyle{\pm}\text{3.47}}$ & \textbf{62.98}$^{\scriptscriptstyle{\pm}\text{8.21}}$ \\
Ours (Betweenness) & \textbf{80.36}$^{\scriptscriptstyle{\pm}\text{3.58}}$ & \textbf{78.24}$^{\scriptscriptstyle{\pm}\text{4.76}}$ & \underline{81.82}$^{\scriptscriptstyle{\pm}\text{1.47}}$ & \underline{81.66}$^{\scriptscriptstyle{\pm}\text{1.61}}$ & 77.97$^{\scriptscriptstyle{\pm}\text{2.41}}$ & 49.19$^{\scriptscriptstyle{\pm}\text{1.07}}$ & \underline{74.18}$^{\scriptscriptstyle{\pm}\text{10.98}}$ & \underline{58.97}$^{\scriptscriptstyle{\pm}\text{9.68}}$ \\
\hline
\hline
\multicolumn{1}{r|}{Dataset} & \multicolumn{4}{c|}{Energy Issues (China) [10.6\%]} & \multicolumn{4}{c}{Insurgency Separatism (Philippines) [2.7\%]}\\
\cline{2-9}
\multicolumn{1}{r|}{Belief Embedding} & \multicolumn{2}{c|}{GCN} & \multicolumn{2}{c|}{SGVGAE} & \multicolumn{2}{c|}{GCN} & \multicolumn{2}{c}{SGVGAE}\\
\cline{2-9}
AL Method & Acc. (\%) & \multicolumn{1}{c|}{Macro F1 (\%)} & Acc. (\%) & Macro F1 (\%) & Acc. (\%) & \multicolumn{1}{c|}{Macro F1 (\%)} & Acc. (\%) & Macro F1 (\%) \\
\hline
Random & 59.84$^{\scriptscriptstyle{\pm}\text{8.78}}$ & 57.44$^{\scriptscriptstyle{\pm}\text{8.31}}$ & 73.67$^{\scriptscriptstyle{\pm}\text{9.72}}$ & 71.47$^{\scriptscriptstyle{\pm}\text{11.16}}$ & 62.16$^{\scriptscriptstyle{\pm}\text{7.68}}$ & 57.36$^{\scriptscriptstyle{\pm}\text{9.21}}$ & 74.96$^{\scriptscriptstyle{\pm}\text{3.90}}$ & 74.77$^{\scriptscriptstyle{\pm}\text{3.83}}$ \\
Centrality (Degree) & 58.86$^{\scriptscriptstyle{\pm}\text{4.28}}$ & 58.02$^{\scriptscriptstyle{\pm}\text{4.73}}$ & 75.33$^{\scriptscriptstyle{\pm}\text{8.55}}$ & 74.47$^{\scriptscriptstyle{\pm}\text{8.38}}$ & 47.05$^{\scriptscriptstyle{\pm}\text{0.81}}$ & 46.53$^{\scriptscriptstyle{\pm}\text{0.99}}$ & 73.67$^{\scriptscriptstyle{\pm}\text{3.22}}$ & 73.39$^{\scriptscriptstyle{\pm}\text{3.17}}$ \\
Centrality (PageRank) & 56.00$^{\scriptscriptstyle{\pm}\text{3.77}}$ & 55.37$^{\scriptscriptstyle{\pm}\text{4.19}}$ & 78.00$^{\scriptscriptstyle{\pm}\text{7.71}}$ & 76.33$^{\scriptscriptstyle{\pm}\text{9.67}}$ & \textbf{66.97}$^{\scriptscriptstyle{\pm}\text{4.15}}$ & \underline{64.59}$^{\scriptscriptstyle{\pm}\text{6.09}}$ & 74.67$^{\scriptscriptstyle{\pm}\text{3.98}}$ & 74.43$^{\scriptscriptstyle{\pm}\text{3.94}}$ \\
Centrality (Betweenness) & 57.33$^{\scriptscriptstyle{\pm}\text{5.59}}$ & 54.69$^{\scriptscriptstyle{\pm}\text{5.33}}$ & 82.67$^{\scriptscriptstyle{\pm}\text{4.13}}$ & 81.91$^{\scriptscriptstyle{\pm}\text{4.20}}$ & 58.40$^{\scriptscriptstyle{\pm}\text{0.75}}$ & 54.93$^{\scriptscriptstyle{\pm}\text{1.07}}$ & 73.33$^{\scriptscriptstyle{\pm}\text{3.55}}$ & 73.16$^{\scriptscriptstyle{\pm}\text{3.53}}$ \\
Entropy & \textbf{64.22}$^{\scriptscriptstyle{\pm}\text{8.54}}$ & 57.98$^{\scriptscriptstyle{\pm}\text{7.21}}$ & 75.33$^{\scriptscriptstyle{\pm}\text{7.78}}$ & 74.25$^{\scriptscriptstyle{\pm}\text{7.75}}$ & 59.47$^{\scriptscriptstyle{\pm}\text{8.99}}$ & 55.62$^{\scriptscriptstyle{\pm}\text{11.57}}$ & 73.88$^{\scriptscriptstyle{\pm}\text{4.45}}$ & 73.69$^{\scriptscriptstyle{\pm}\text{4.38}}$ \\
AGE & 52.27$^{\scriptscriptstyle{\pm}\text{10.99}}$ & 47.90$^{\scriptscriptstyle{\pm}\text{9.95}}$ & 78.00$^{\scriptscriptstyle{\pm}\text{8.94}}$ & 76.11$^{\scriptscriptstyle{\pm}\text{10.94}}$ & 56.55$^{\scriptscriptstyle{\pm}\text{2.15}}$ & 52.35$^{\scriptscriptstyle{\pm}\text{3.30}}$ & 73.56$^{\scriptscriptstyle{\pm}\text{4.80}}$ & 73.37$^{\scriptscriptstyle{\pm}\text{4.72}}$ \\
ANRMAB & 57.44$^{\scriptscriptstyle{\pm}\text{9.96}}$ & 54.21$^{\scriptscriptstyle{\pm}\text{9.68}}$ & 78.33$^{\scriptscriptstyle{\pm}\text{9.41}}$ & 76.48$^{\scriptscriptstyle{\pm}\text{11.41}}$ & 56.76$^{\scriptscriptstyle{\pm}\text{4.22}}$ & 51.75$^{\scriptscriptstyle{\pm}\text{3.75}}$ & 72.15$^{\scriptscriptstyle{\pm}\text{5.23}}$ & 71.98$^{\scriptscriptstyle{\pm}\text{5.15}}$ \\
GPA & 59.76$^{\scriptscriptstyle{\pm}\text{6.84}}$ & \underline{59.03}$^{\scriptscriptstyle{\pm}\text{6.74}}$ & 78.67$^{\scriptscriptstyle{\pm}\text{8.75}}$ & 76.97$^{\scriptscriptstyle{\pm}\text{10.45}}$ & 57.98$^{\scriptscriptstyle{\pm}\text{2.47}}$ & 55.68$^{\scriptscriptstyle{\pm}\text{3.30}}$ & 74.00$^{\scriptscriptstyle{\pm}\text{5.06}}$ & 73.80$^{\scriptscriptstyle{\pm}\text{4.99}}$ \\
RIM & 63.14$^{\scriptscriptstyle{\pm}\text{5.01}}$ & 56.49$^{\scriptscriptstyle{\pm}\text{3.80}}$ & 78.67$^{\scriptscriptstyle{\pm}\text{5.47}}$ & 77.69$^{\scriptscriptstyle{\pm}\text{5.77}}$ & 54.67$^{\scriptscriptstyle{\pm}\text{0.00}}$ & 37.87$^{\scriptscriptstyle{\pm}\text{0.00}}$ & 73.89$^{\scriptscriptstyle{\pm}\text{3.39}}$ & 73.64$^{\scriptscriptstyle{\pm}\text{3.39}}$ \\ \hdashline
Ours (PageRank) & 58.86$^{\scriptscriptstyle{\pm}\text{8.55}}$ & 53.74$^{\scriptscriptstyle{\pm}\text{7.85}}$ & \underline{82.67}$^{\scriptscriptstyle{\pm}\text{2.31}}$ & \underline{82.19}$^{\scriptscriptstyle{\pm}\text{2.36}}$ & 60.57$^{\scriptscriptstyle{\pm}\text{3.34}}$ & 58.99$^{\scriptscriptstyle{\pm}\text{4.07}}$ & \underline{77.14}$^{\scriptscriptstyle{\pm}\text{3.12}}$ & \underline{76.86}$^{\scriptscriptstyle{\pm}\text{3.06}}$ \\
Ours (Betweenness) & \underline{60.57}$^{\scriptscriptstyle{\pm}\text{14.86}}$ & \textbf{59.43}$^{\scriptscriptstyle{\pm}\text{14.81}}$ & \textbf{83.00}$^{\scriptscriptstyle{\pm}\text{3.83}}$ & \textbf{82.43}$^{\scriptscriptstyle{\pm}\text{4.14}}$ & \underline{65.33}$^{\scriptscriptstyle{\pm}\text{1.89}}$ & \textbf{64.78}$^{\scriptscriptstyle{\pm}\text{1.81}}$ & \textbf{78.33}$^{\scriptscriptstyle{\pm}\text{1.28}}$ & \textbf{78.06}$^{\scriptscriptstyle{\pm}\text{1.25}}$ \\
\hline
\end{tabular}
\end{table*}

\subsection{Experimental Setup}
\subsubsection{Datasets}

We employ six real-world datasets collected from the X platform using a combination of keyword and hashtag filters. These datasets contain social media posts which are processed by identifying re-posts (previously called retweets) and merging them into unified "assertions" for the belief representation learning task. Below is a brief description of each dataset, along with the labels used (note: neutral labels are not discussed in this section).

\begin{itemize}
    \item \textbf{Eurovision 2016}: This dataset was collected during the Eurovision Song Contest in 2016, focusing primarily on the competition and the Ukrainian singer Susana Jamaladinova, known professionally as Jamala. The labels are either pro-Jamala or anti-Jamala. Pro-Jamala users expressed support for her victory, while anti-Jamala users claimed the contest results were politically influenced.
    \item \textbf{Russia/Ukraine Conflict (Visual)}: Collected between May and November 2022, this dataset was created using pro-Russia and pro-Ukraine keyword filters. Example keywords include ``\#standwithputin'', ``Minsk Accord'', ``NATO'', and ``BRICS''. This dataset consists of only visual assertions, which are the clustered media (images) attached under the social media posts using CLIP~\cite{radford2021learning}. The labels are pro-Ukraine or pro-Russia.
    \item \textbf{EDCA (Philippines), Territorial Sovereignty (Philippines), Insurgency Separatism (Philippines), and Energy Issues (China)}: Collected in 2023 using keyword filters related to geopolitical topics involving the Philippines, the United States, China, and the South China Sea. The dataset is further segmented into distinct topics using the PIEClass model~\cite{zhang-etal-2023-pieclass}, a weakly supervised topic classification model that employs topic-specific seed keywords.
    \begin{itemize}
        \item \textbf{EDCA (Philippines)}: Related keywords are ``Enhanced Defense Coorperation Agreement'', ``EDCA'', and etc. Labels are pro- or anti-EDCA.
        \item \textbf{Territorial Sovereignty (Philippines)}: Related keywords are ``territorial sovereignty'' and etc. Labels are pro- or anti-West.
        \item \textbf{Energy Issues (China)}: Related keywords are ``energy'', ``manufacturing'', and etc. Labels are pro- or anti-China.
        \item \textbf{Insurgency Separatism (Philippines)}: Related keywords are ``Separatism'', ``autonomy'', and etc. Labels are pro- or anti-Philippines.
    \end{itemize}
\end{itemize}

Table~\ref{table:datasets} provides a summary of the statistics for these datasets.

\begin{figure*}[h]
  \centering
  \begin{subfigure}[b]{\textwidth}
      \includegraphics[width=\textwidth]{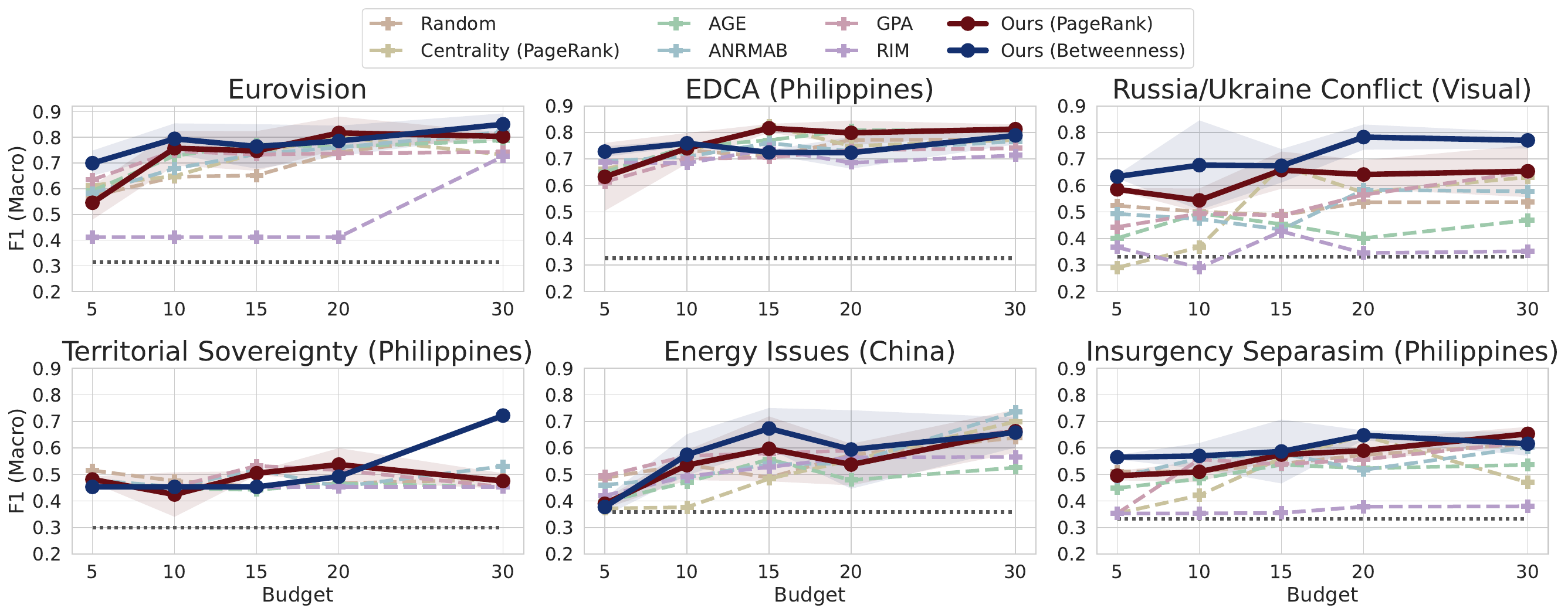}
      \caption{Belief representation learning using GCN.}
      \label{fig:budget_gcn}
  \end{subfigure}
  \begin{subfigure}[b]{\textwidth}
      \includegraphics[width=\textwidth]{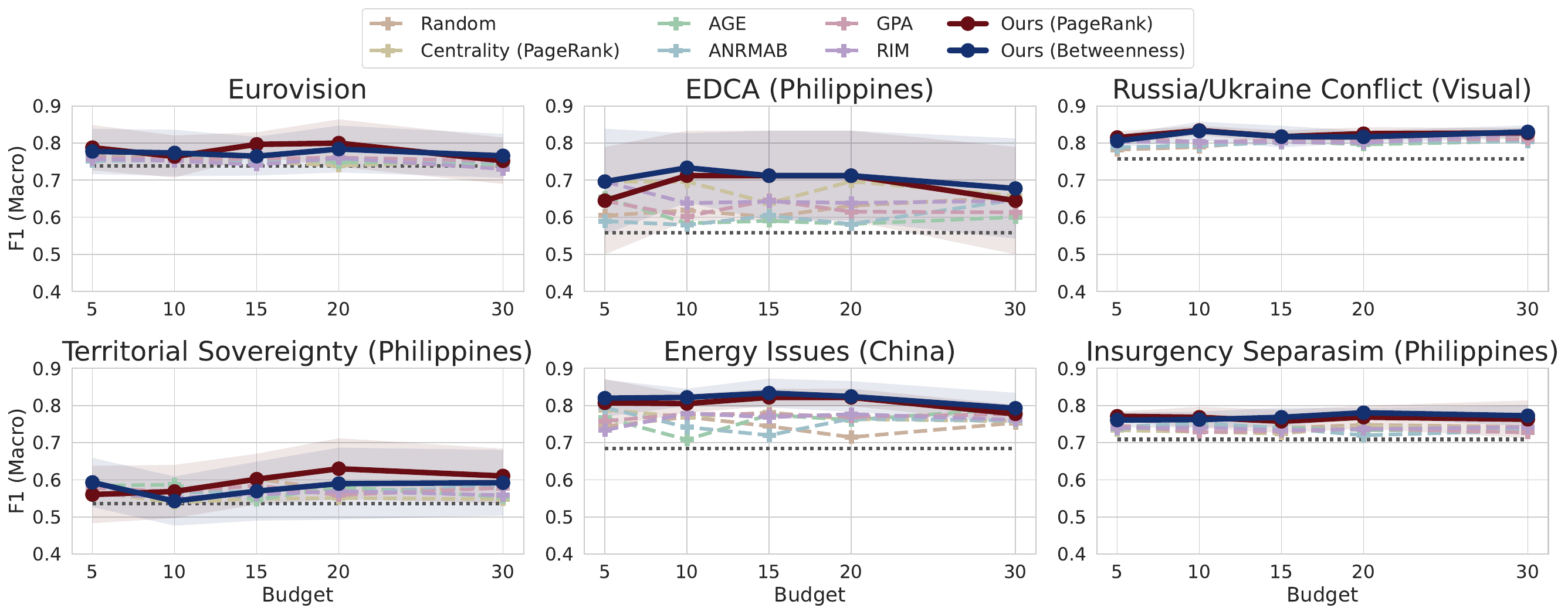}
      \caption{Belief representation learning using SGVGAE.}
      \label{fig:budget_infovgae}
  \end{subfigure}
  \caption{Macro F1 score curve of belief representation models trained in a semi-supervised fashion on AL-queried nodes under different budget constraints. The dotted flat gray line represents the performance of the model trained in an unsupervised fashion.}\label{fig:series}
\end{figure*}

\subsubsection{Baselines}
We compare our method against the following baseline methods:
\begin{itemize}
    \item \textbf{Random}: This approach selects $B$ nodes randomly for annotation.
    \item \textbf{Centrality-based}: This method selects $B$ nodes for annotation based on a specific centrality metric (e.g., degree, PageRank, or betweenness).
    \item \textbf{Entropy-based}: At each step, this method selects the node with the highest information entropy in its predicted label distribution, as generated by the classification Graph Neural Network (GNN).
    \item \textbf{AGE}~\cite{cai2017active}: This method selects the node with the highest score at each step, where the score is a step-sensitive combination of three factors: the node's label distribution entropy, a centrality metric, and its representativeness (i.e., the distance of the node's embedding from the nearest cluster center).
    \item \textbf{ANRMAB}~\cite{anrmab}: This approach uses the same three heuristic factors as AGE but combines them dynamically via a multi-armed bandit framework to calculate the node's score.
    \item \textbf{GPA}~\cite{hu2020graph}: In this method, active learning is framed as a Reinforcement Learning (RL) problem. An agent is trained to select nodes based on the current graph state, which is implicitly modeled by a classification GNN.
    \item \textbf{RIM}~\cite{zhang2021rim}: This method uses influence functions to estimate the impact of selected samples and scales the influence-based score by label reliability. Active learning is formulated as a reliable influence maximization problem.
\end{itemize}

\subsubsection{Metrics and Hyperparameters}

We assess the quality of our queried nodes with other active learning methods by evaluating the belief classification task using accuracy and macro F1 score. We employ two belief representation learning models, the Graph Convolutional Network (GCN) and a specialized model called SGVGAE~\cite{li2024large}. In evaluating the belief classification, we focus on messages with annotations. User preferences can be inferred by aggregating their message history, but this may compound our goal of assessing the quality of the chosen active learning nodes. Therefore, reported accuracy and macro F1 score are only for messages.

Experiments were repeated ten times with three train-validation splits, and we report the mean and standard deviation. The following are for the hyperparameters. We have two variants of the \model which employ different centrality metrics. The default number of perturbed graphs is ten ($a_e=4, a_m=3, a_p=3$). For five perturbed graphs, the configuration is (2, 2, 1), and for fifteen perturbed graphs, the configuration is (6, 5, 4).

GCNs used for belief representation learning and the learning model within the model comprises two Graph Convolutional Network (GCN) layers with 32 hidden dimensions and 16 output dimensions. We trained the GNN models using the Adam optimizer with a learning rate of 0.02 and a weight decay of 0.005. For the Stochastic Graph Variational Graph Auto-Encoder (SGVGAE) model, we initialized it with 32 hidden dimensions and it produced two-dimensional embeddings. We trained the SGVGAE using the Adam optimizer with a learning rate of 0.2 and a weight decay of 0.0001.

\subsection{Evaluation Results}

\subsubsection{Belief Representation Learning with A Budget of 20 Queries}

The detailed evaluation results are presented in Table~\ref{table:eval}. Our method has improved belief classification performance across all six datasets. The average improvements in macro F1 score for both GCN and SGVGAE are 2.60\% and 2.51\%, respectively. Furthermore, the accuracy score for SGVGAE shows an average improvement of 3.49\%. Additionally, our method consistently achieves the second-highest performance, which demonstrates its effectiveness.

It was also observed that when baseline methods outperform ours, they typically utilize graph structural statistics and heuristics (centrality and AGE). Although we also employ these measures, we leverage them in a different manner, which will be further discussed in the ablation section. Conversely, GPA and RIM methods are not leading. One possible reason for this is that our dataset contains two types of nodes. While these methods allow for masking when selecting the next query, they do not adapt as effectively as statistical and heuristic methods.

\subsubsection{Belief Representation Learning with Other Allocated Budget Amounts}

We also investigate the performance of our model under different budget constraints. The corresponding plots can be found in Figure~\ref{fig:series}. The dark lines in the plots represent the two variants mentioned in the table. It is evident from the plot that our method consistently outperforms other baselines.
We observe that SGVGAE efficiently utilizes labeled data and maintains consistent performance, irrespective of the budget. Conversely, GCN demonstrates noticeable improvement as the budget increases.

\subsubsection{Examination and Comparison of Queried Nodes}

\begin{figure}[htbp]
    \centering
    \begin{subfigure}{0.49\columnwidth}
        \centering
        \includegraphics[width=\linewidth]{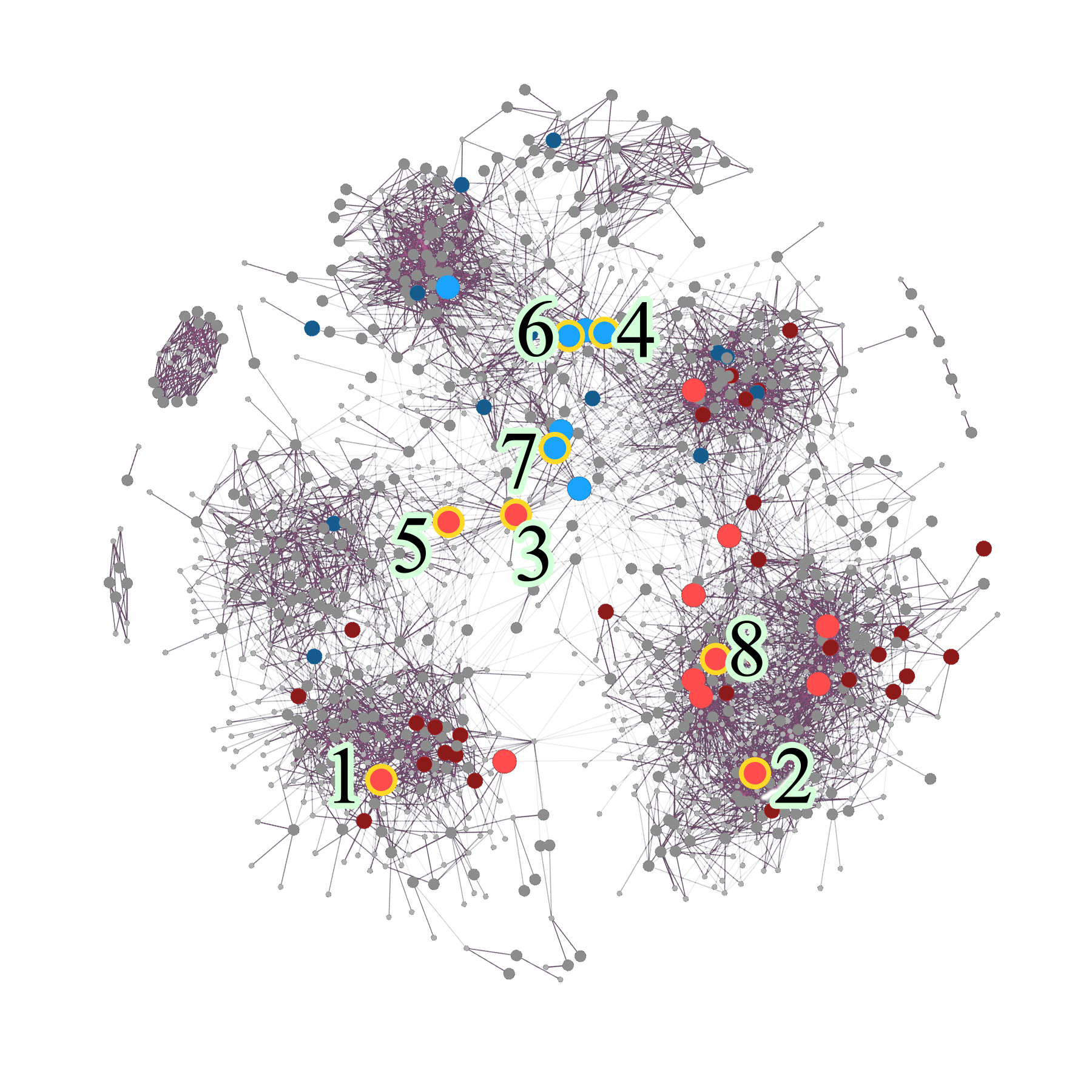}
        \caption{Centrality (Betweenness)}
    \end{subfigure}
    \hfill
    \begin{subfigure}{0.49\columnwidth}
        \centering
        \includegraphics[width=\linewidth]{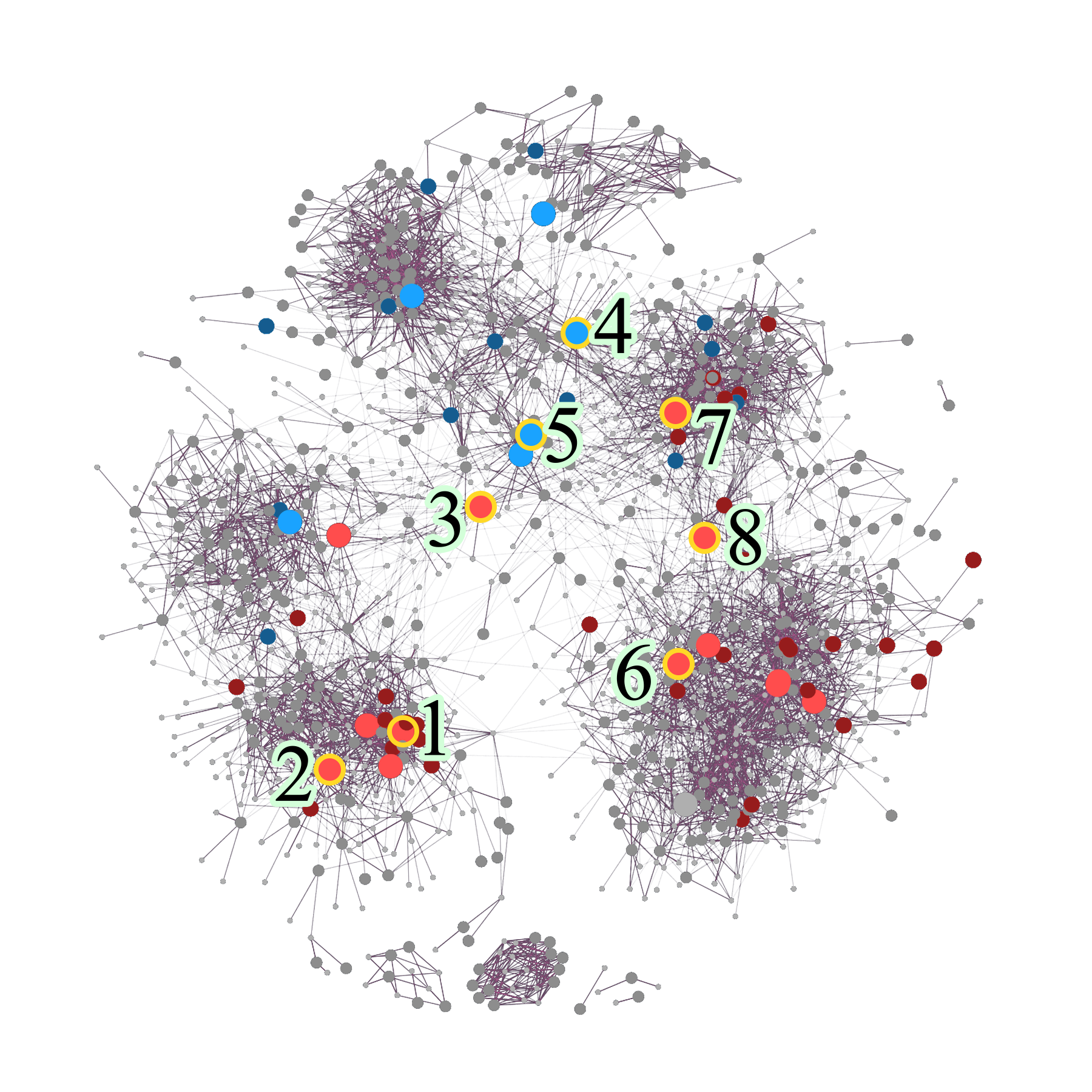}
        \caption{AGE}
    \end{subfigure}
    \begin{subfigure}{0.49\columnwidth}
        \centering
        \includegraphics[width=\linewidth]{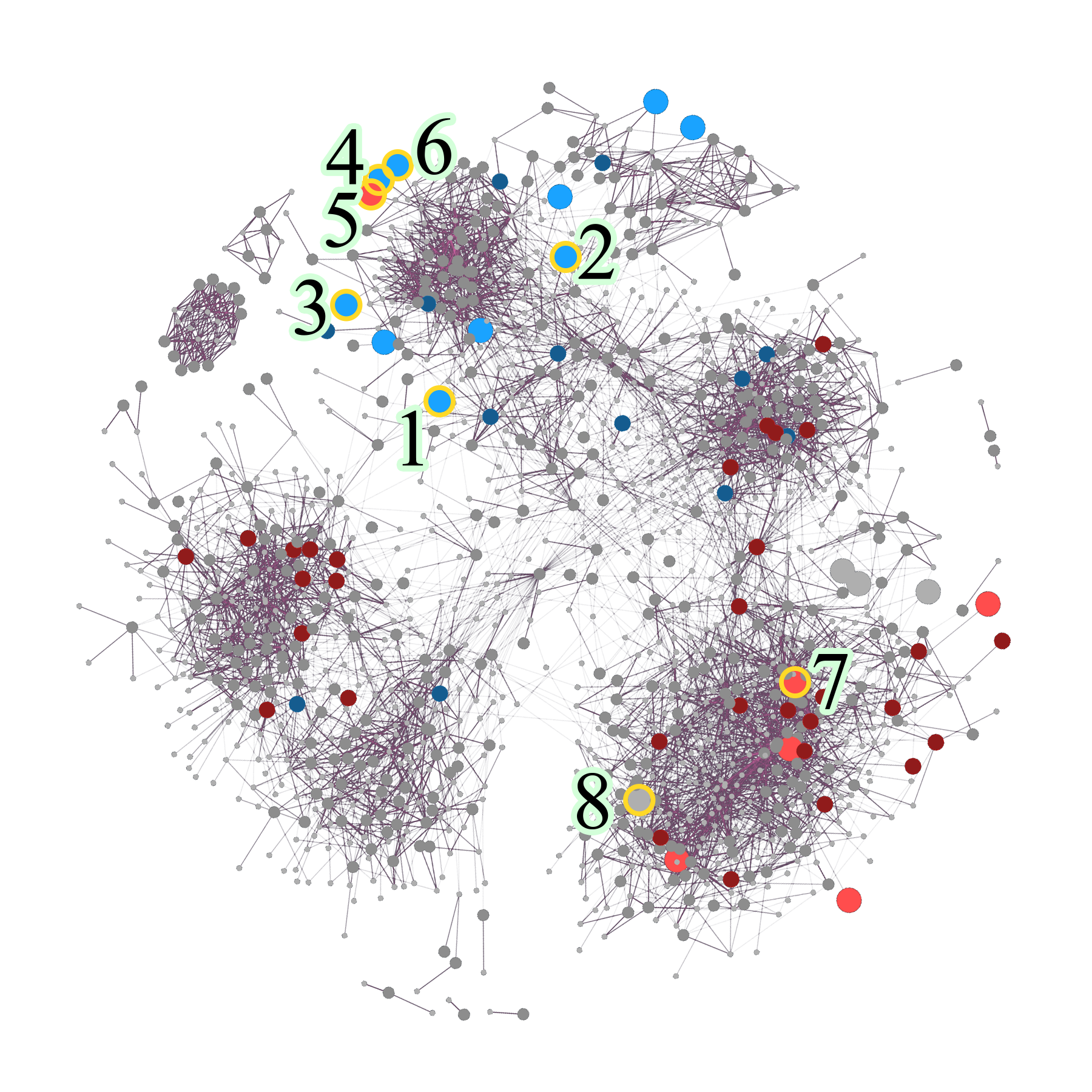}
        \caption{GPA}
    \end{subfigure}
    \hfill
    \begin{subfigure}{0.49\columnwidth}
        \centering
        \includegraphics[width=\linewidth]{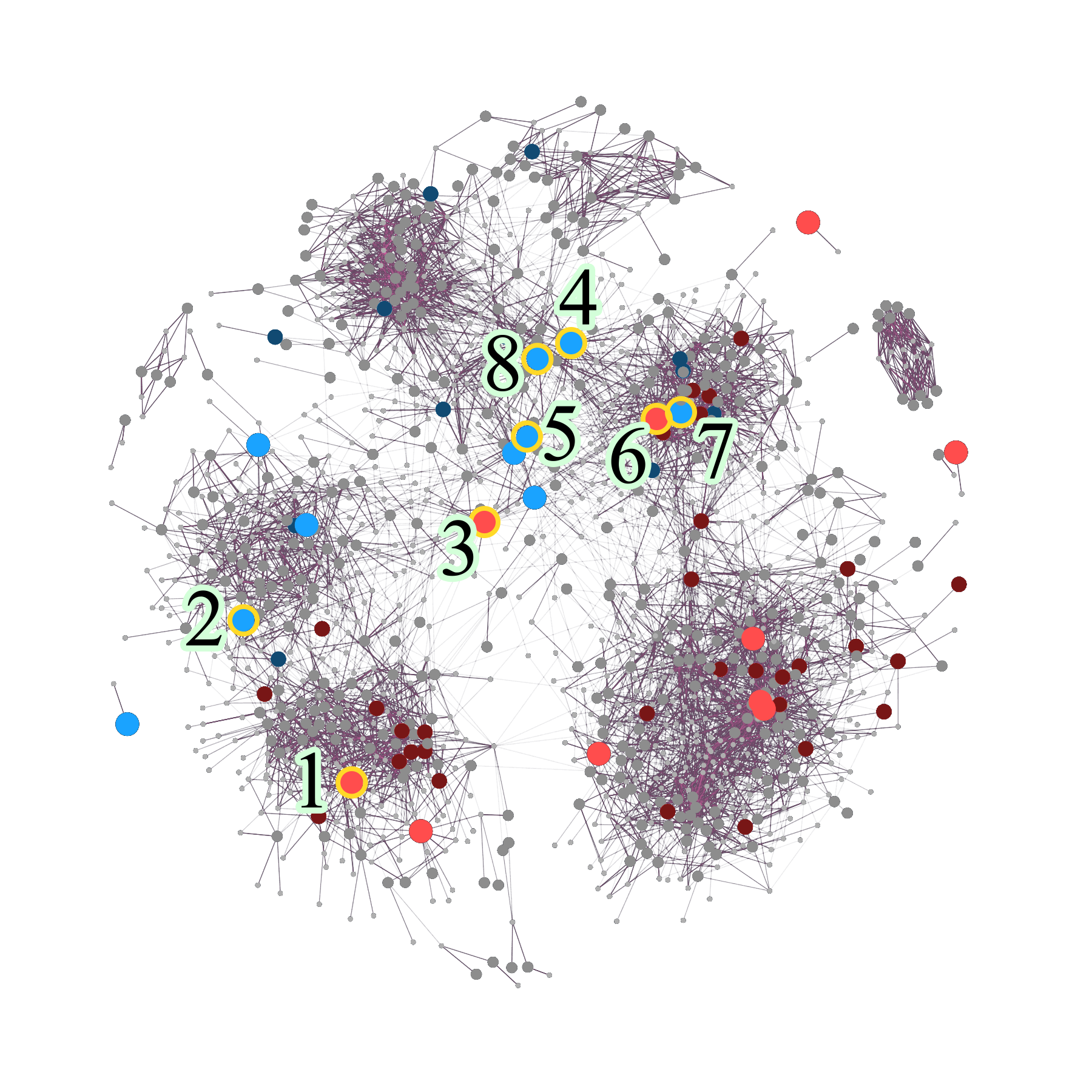}
        \caption{\model (PageRank)}
    \end{subfigure}

    \caption{Visualization of EDCA dataset with 20 queried and testing nodes}
    \label{fig:examine_edca}
\end{figure}

\begin{figure}[htbp]
    \centering
    \begin{subfigure}{0.49\columnwidth}
        \centering
        \includegraphics[width=\linewidth]{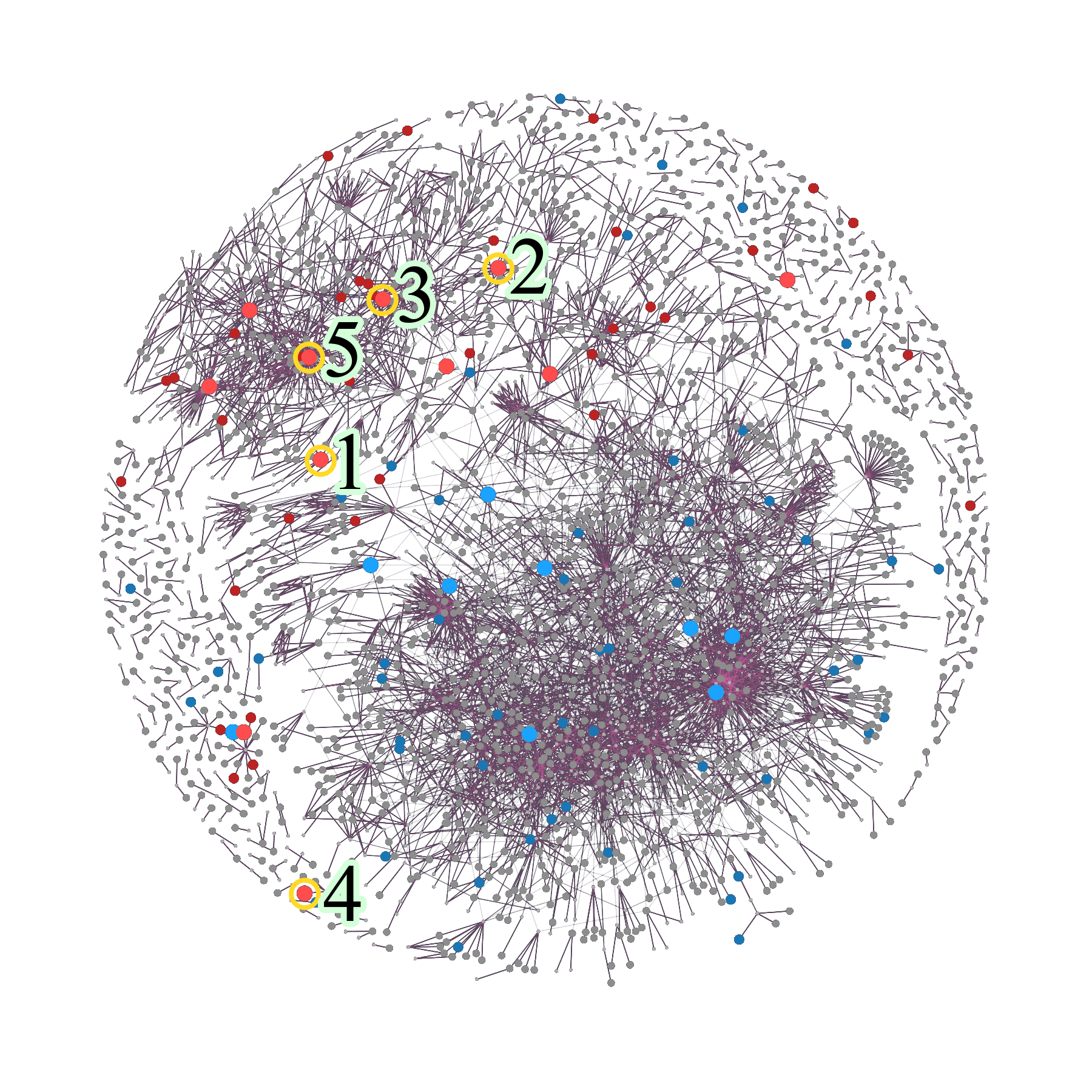}
        \caption{Centrality (Betweenness)}
    \end{subfigure}
    \hfill
    \begin{subfigure}{0.49\columnwidth}
        \centering
        \includegraphics[width=\linewidth]{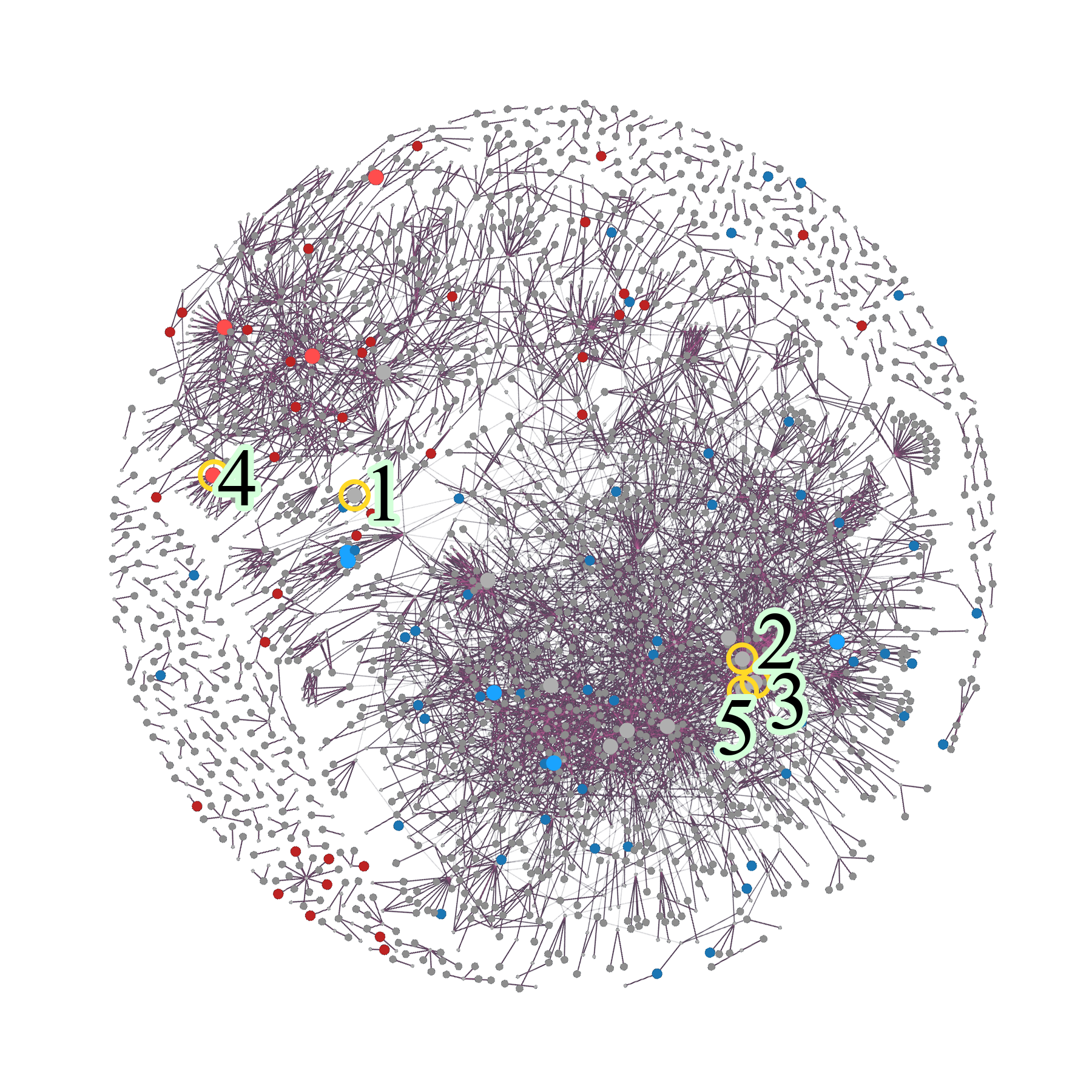}
        \caption{AGE}
    \end{subfigure}
    \begin{subfigure}{0.49\columnwidth}
        \centering
        \includegraphics[width=\linewidth]{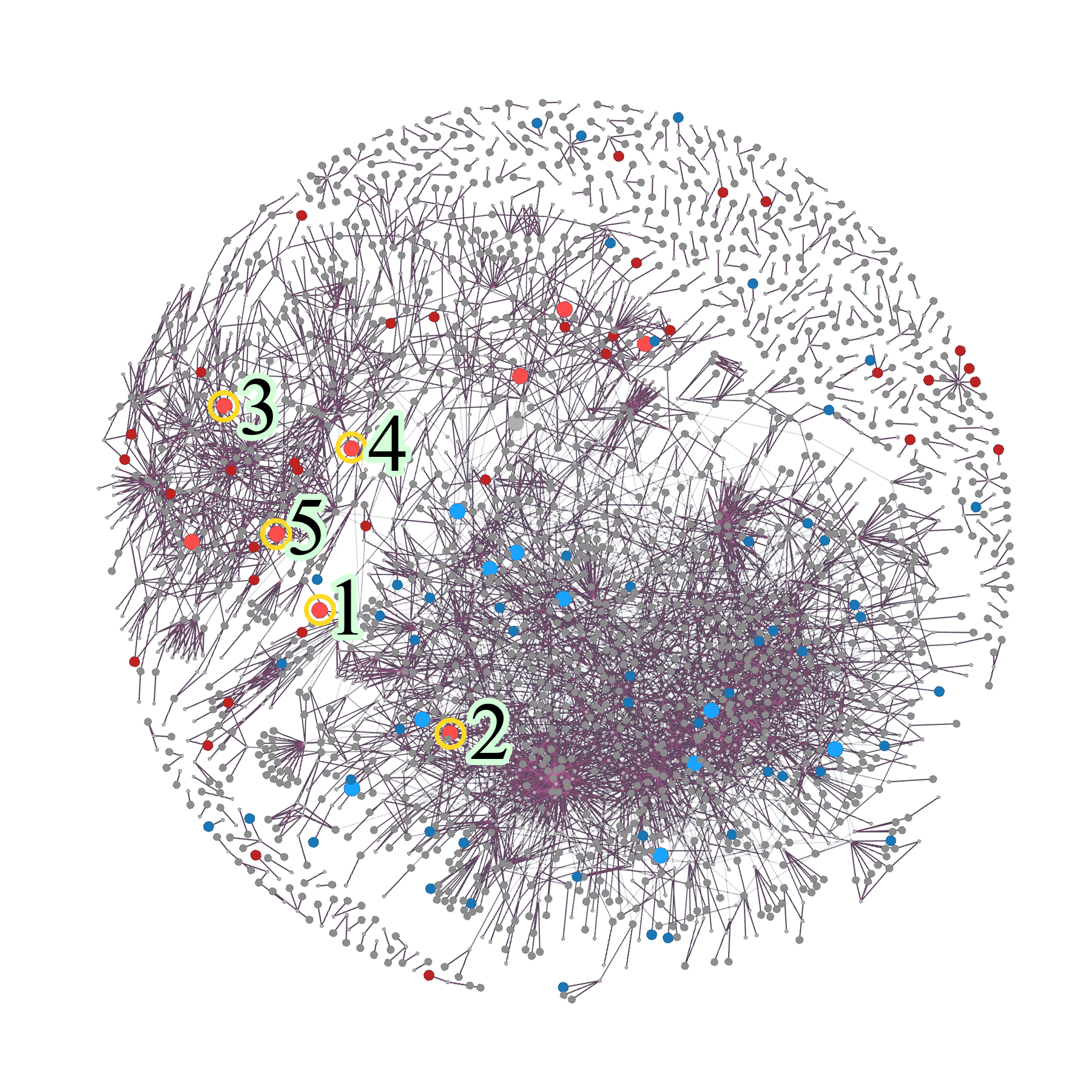}
        \caption{GPA}
    \end{subfigure}
    \hfill
    \begin{subfigure}{0.49\columnwidth}
        \centering
        \includegraphics[width=\linewidth]{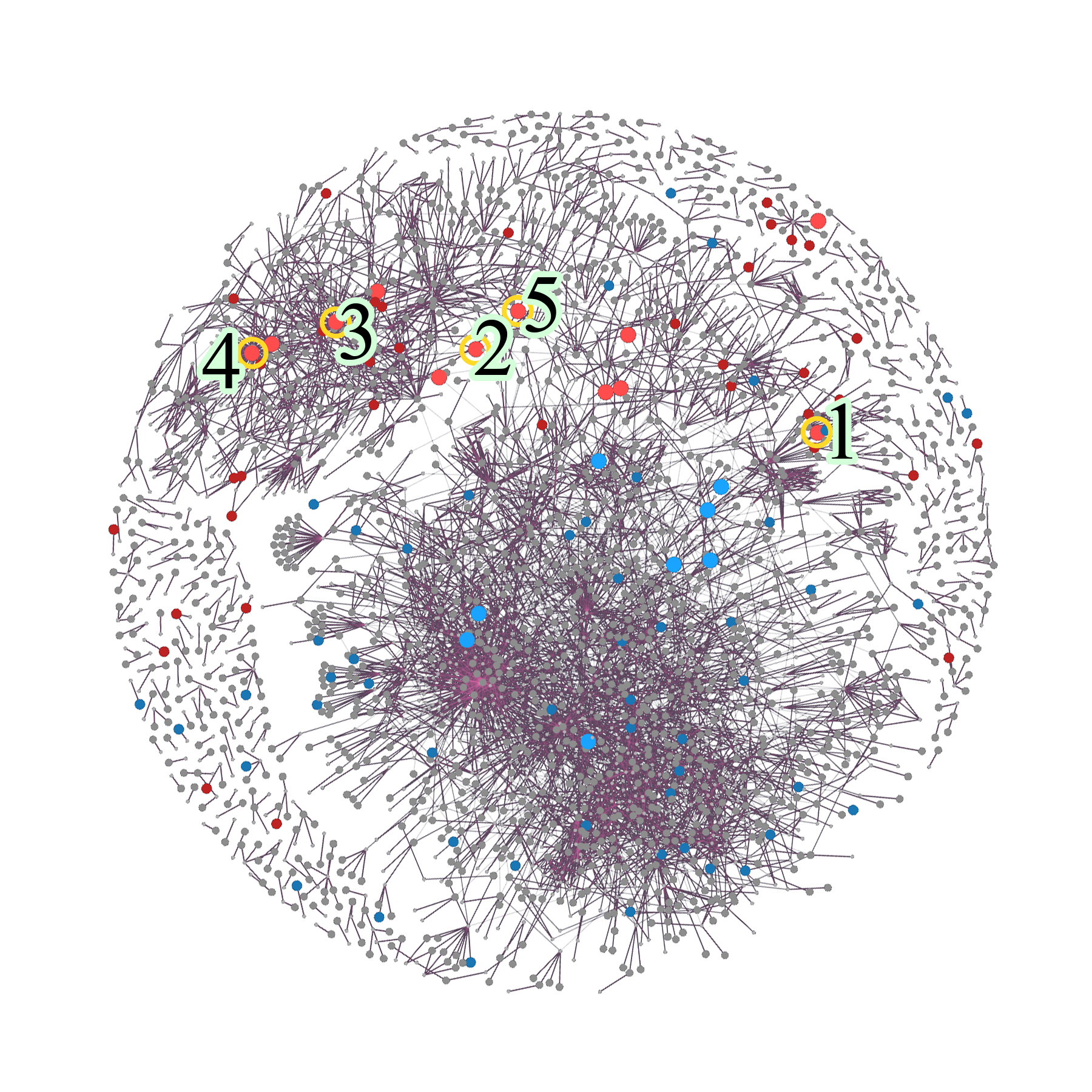}
        \caption{\model (PageRank)}
    \end{subfigure}

    \caption{Visualization of Russia/Ukraine Conflict (Visual) dataset with 20 queried nodes and testing nodes}
    \label{fig:examine_vis}
\end{figure}

To further validate the capabilities of \model, we visualized the queried nodes in the EDCA and Russia/Ukraine Conflict datasets using a budget of 20 in a graph visualization tool called Cosmograph. The results can be seen in Figure~\ref{fig:examine_edca} and Figure~\ref{fig:examine_vis}. We have provided clear labels indicating (1) the order of the queries up to the eighth node, (2) the query labels marked in red and blue, and (3) the nodes and their labels in the test set shown in darker red and blue.

In Figure~\ref{fig:examine_edca}, we observed that our method efficiently utilizes the budget by selecting nodes 1 and 2, which effectively delineates two clusters. Additionally, the selection of nodes 6 and 7 is crucial, as they represent conflicting yet interconnected nodes. Our method appears to prioritize labeling nodes at the intersection of clusters while also allowing for the labeling of isolated nodes, provided that the budget permits.

\begin{figure}[h]
  \centering
  \includegraphics[width=.98\linewidth]{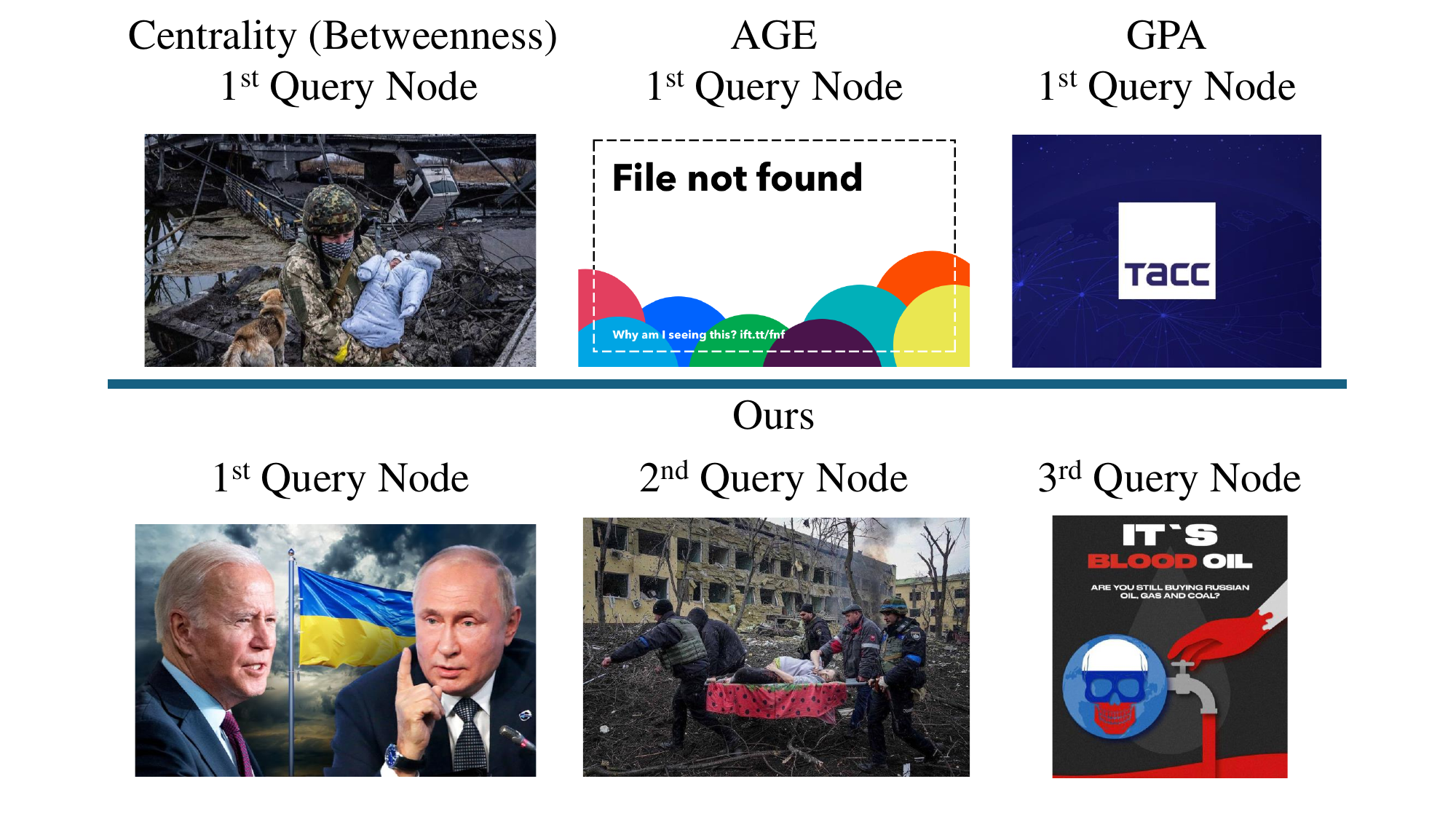}
  \caption{Comparison of queried visual assertions in the Russia/Ukraine Conflict dataset by different AL methods. Results are corresponding to the graph visualization in Figure~\ref{fig:examine_vis}.}
  \label{fig:examine_vis_sample}
\end{figure}

In Figure~\ref{fig:examine_vis}, we observed that all methods, except AGE, select nodes around the primary cluster, which seems to be the boundary as well. Meanwhile, only we and centrality-based methods selected a meaningful first query, as demonstrated in Figure~\ref{fig:examine_vis_sample}

\subsection{Ablation Study}
\begin{figure}[h]
  \centering
  \includegraphics[width=\linewidth]{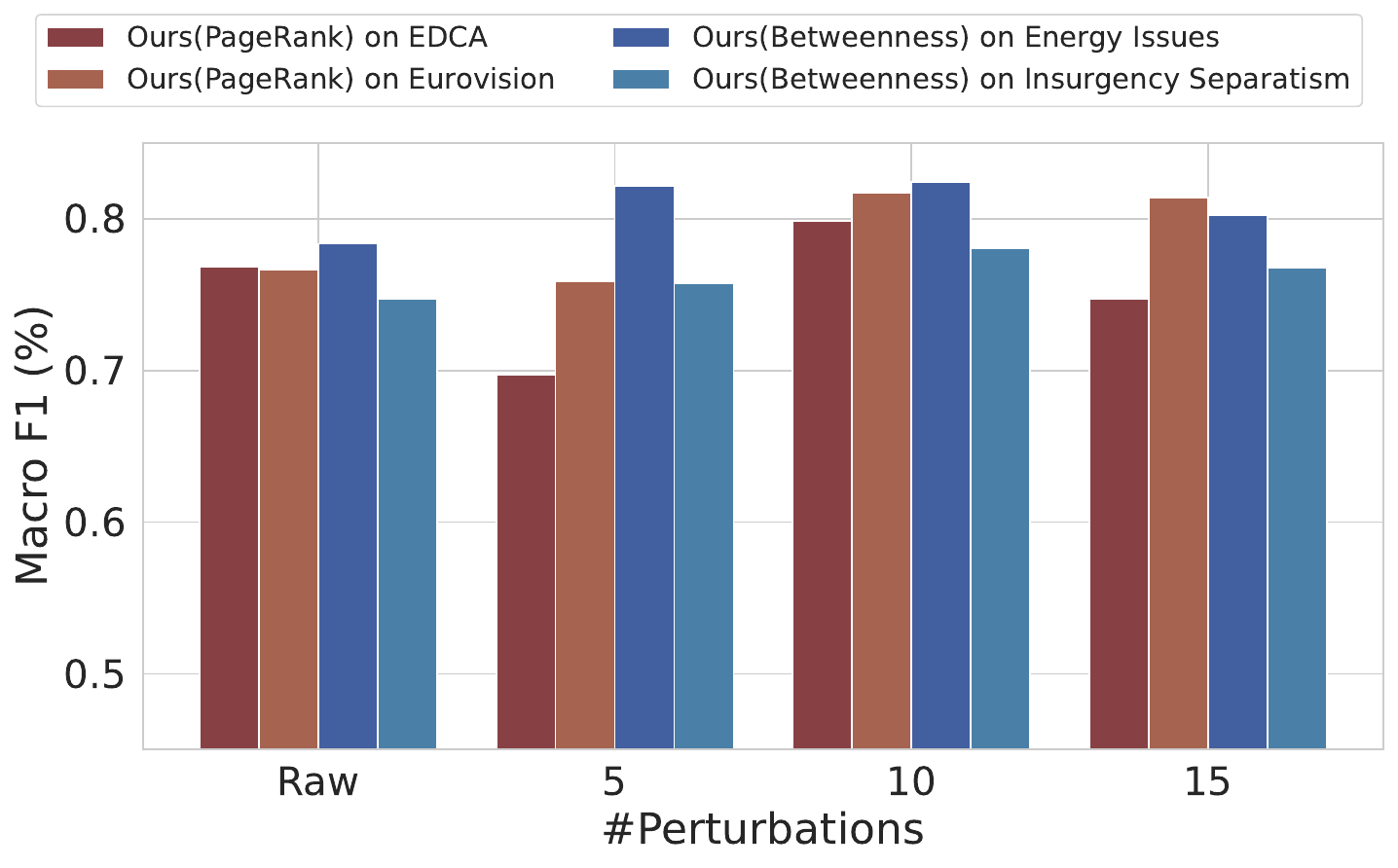}
  \caption{The Macro F1 score of believe representation learning with four variants of \model.}
  \label{fig:ablation_perturb}
\end{figure}

In this study, we aim to explore how the performance of the \model is impacted by the number of perturbed graphs. Additionally, we seek to investigate the distinction between sensitivity scores and direct centrality metrics. Our experiments encompass all datasets across four configurations: utilizing raw centrality metrics, and incorporating five, ten, or fifteen perturbed graphs. The results of our findings are illustrated in Figure~\ref{fig:ablation_perturb}. ``Raw (10)'' means that the sensitivity module is replaced by direct centrality metric, while the instability score still uses 10 perturbed graphs.

The sensitivity score is indeed different from direct metrics since the performance comparison to the third group is clear. When we further compare the groups using five or fifteen perturbations to the third group, they also do not perform well. We hypothesize that having too few perturbed graphs may result in a high dynamic range of the sensitivity score due to extreme cases while having too many could result in a reduced dynamic range due to smoothing and the law of large numbers. Therefore, the number of perturbations is a hyperparameter of \model, and in this paper, we use 10.

\subsection{Computational Costs}

\begin{figure}[h]
  \centering
  \begin{subfigure}[b]{\linewidth}
      \includegraphics[width=\linewidth]{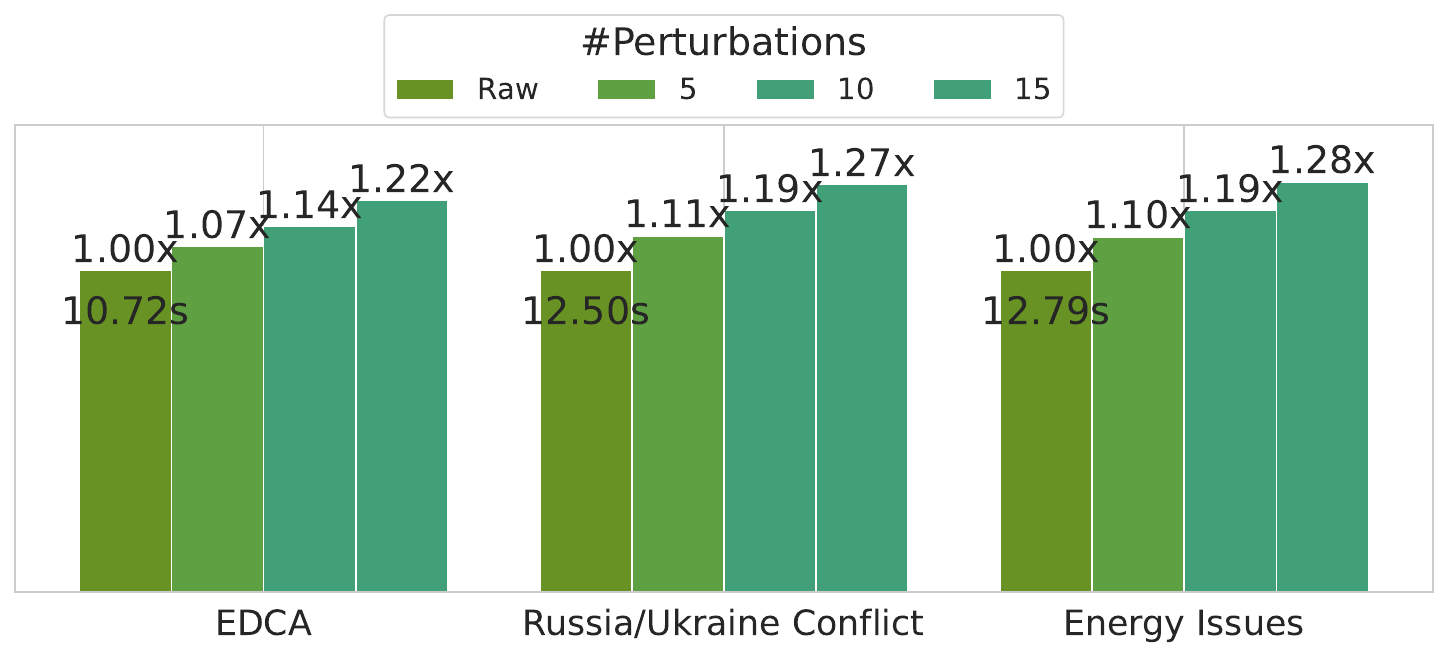}
      \caption{Using PageRank as the sensitivity score.}
      \label{fig:pagerank_complex}
  \end{subfigure}
  \begin{subfigure}[b]{\linewidth}
      \includegraphics[width=\linewidth]{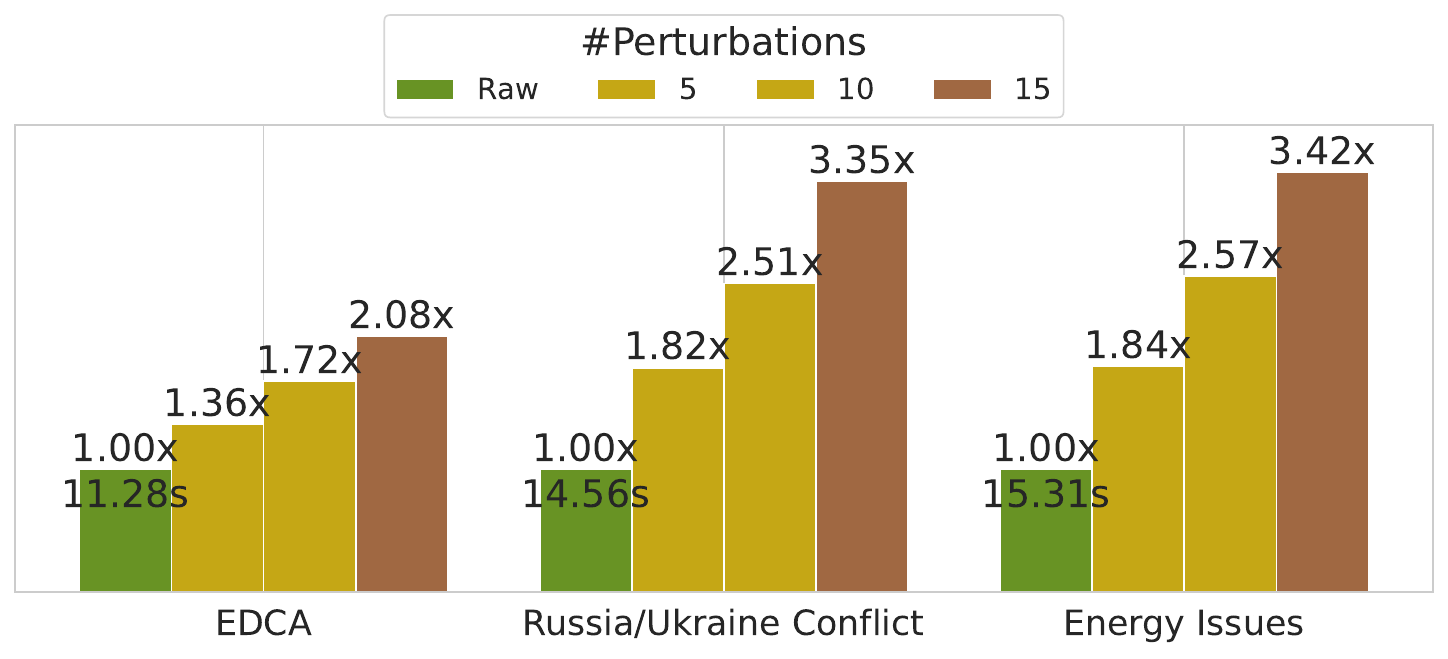}
      \caption{Using betweenness centrality as the sensitivity score.}
      \label{fig:betweenness_complex}
  \end{subfigure}
  \caption{Increase in total time elapsed for \model to complete the node queries with a budget of 20}
  \label{fig:time_complex}
\end{figure}

Adding perturbed graphs that also require graph neural network encoding and centrality metric computation will incur costs. However, these costs are usually one-time expenses or considered acceptable compared to the downstream tasks, which will likely take much longer to work with the selected nodes. We conducted a study using an extreme example where the downstream task takes a similar amount of time to finish as the active learning algorithm. We then used a version that does not calculate centrality metrics on every perturbed graph as a baseline to measure the running time difference as the active learning method uses more perturbed graphs. The results are presented in Figure~\ref{fig:time_complex}. In it, we used three datasets with increasing edge counts to illustrate the trend when working with larger datasets. It's evident that the running time of PageRank centrality increases at a fairly acceptable rate, while the running time of betweenness centrality grows rapidly. We recognize this phenomenon and recommend the use of PageRank centrality on larger datasets. Additionally, in the future, \model can be expanded to calculate a local estimate version of the centrality metrics to address this issue.

\section{Related Works}\label{sec:related}
\subsection{Graph Neural Networks} Graph Neural Networks (GNNs) have achieved significant success by learning to aggregate information from neighboring nodes in a graph, thereby making more reliable predictions. Common GNN architectures include Graph Convolutional Networks (GCNs)\cite{Kipf:2016tc} and Graph Attention Networks (GATs)\cite{velivckovic2017graph}. GNNs have enabled various important machine learning tasks, such as (i) node classification, which aims to predict the label of each node in a graph based on its features and connectivity~\cite{li2022sentiment,zarandi2024survey,zhang2024doubleh}, and (ii) graph representation learning~\cite{kipf2016variational}, which focuses on learning low-dimensional representations for nodes that capture their structural and semantic information.

\subsection{Semi-Supervised Graph Representation Learning} Graph representation learning~\cite{chen2020graph} is a fundamental problem in graph-based machine learning, which seeks to learn node representations through training objectives designed to preserve graph structure and properties. This has been applied to various tasks, including clustering-based community detection in social networks~\cite{bruna2017community,jiang2021social,fortunato2010community}, graph visualization~\cite{van2008visualizing,huang2020gnnvis}, and downstream supervised tasks~\cite{xiao2020timme,ying2018graph}. For example, Deep Graph Infomax (DGI)\cite{velickovic2019deep} proposes a self-supervised objective based on maximizing mutual information between local and global graph representations. Variational Graph Autoencoders (VGAEs)\cite{kipf2016variational} introduce a variational auto-encoder framework that learns to reconstruct graph structure and features via an encoder-decoder architecture.

Unsupervised~\cite{li2022unsupervised,kipf2016variational} and semi-supervised~\cite{song2022graph,kipf2016semi,li2024large} graph learning models have proven to be particularly effective and robust when available annotations are limited. Existing research has explored graph representation learning through approaches such as collaborative filtering~\cite{yang2014survey,pham2011clustering}, non-negative matrix factorization~\cite{al2017unveiling}, random walks~\cite{perozzi2014deepwalk,tang2015line}, or GCNs~\cite{li2018deeper}. More recent work has investigated graph learning models based on graph auto-encoders~\cite{kipf2016variational}. For instance, InfoVGAE~\cite{li2022unsupervised} proposes an unsupervised objective to create a disentangled non-negative embedding space that preserves orthogonality and interpretability. The ARVGA model~\cite{pan2019learning} introduces an adversarial training objective to enhance robustness for graphs with outliers. The SGVGAE model~\cite{li2024large} proposes a semi-supervised training objective combining the disentangled embeddings with large language model-generated soft annotations to improve the performance on sparse or noisy graph data. However, most existing semi-supervised models rely on random strategy or simple degree-based heuristics for selecting node annotations for semi-supervision, which limits the effectiveness of learned node representations.

\subsection{Active Learning} Active learning~\cite{settles2009active} has been extensively explored in i.i.d. data across various domains, including natural language processing~\cite{shen-etal-2017-deep} and computer vision~\cite{9}. Recently, there has been growing interest in applying active learning to graph-structured data. Early approaches~\cite{10.1145/2623330.2623760,KDD-2013-GuALH} were built on the graph homophily assumption, which suggests that neighboring nodes are more likely to share the same label, and leveraged graph signal processing theories to select nodes for active learning. In contrast, our proposed method does not rely on the strict homophily assumption, making it applicable to a broader range of real-world social graphs.

Recent studies have leveraged the power of GNNs to design more effective selection criteria. AGE~\cite{cai2017active} estimates node informativeness using a linear combination of three heuristics, with weights sampled from a time-sensitive beta distribution. ANRMAB~\cite{anrmab} also combines heuristics but dynamically adjusts the weights via a multi-armed bandit approach, while ActiveHNE~\cite{chen2019activehneactiveheterogeneousnetwork} applies a similar framework to heterogeneous graphs. In contrast, \model is based on graph perturbation and selects nodes by measuring performance variances, avoiding predefined heuristics and improving generalization across diverse social graphs with multi-modal nodes, such as text and images.

\section{Conclusion}\label{sec:conclusion}
We addressed the challenge of optimizing labeling resource allocation for semi-supervised belief representation learning in social networks. We introduced a novel perturbation-based active learning strategy, \model, inspired by graph data augmentation, which automatically selects nodes for labeling without requiring human intervention. The core idea is that messages in the network with heightened sensitivity to structural variations signal key nodes that can significantly impact the weak-supervision process. To capture this, we designed an automatic estimator that selects nodes based on prediction variance across various graph perturbations, ensuring that the method remains model-agnostic and application-independent. Extensive experiments demonstrated the effectiveness of our approach, highlighting its potential for improving belief representation learning tasks in diverse social network scenarios.

\section*{Acknowledgment}

Research reported in this paper was sponsored in part by the DARPA award HR001121C0165, the DARPA award HR00112290105, the DoD Basic Research Office award HQ00342110002, the Army Research Laboratory under Cooperative Agreement W911NF-17-20196. It was also partially supported by ACE, one of the seven centers in JUMP 2.0, a Semiconductor Research Corporation (SRC) program sponsored by DARPA.

\bibliographystyle{IEEEtran}
\bibliography{ref}

\begin{thebibliography}{10}
\providecommand{\url}[1]{#1}
\csname url@samestyle\endcsname
\providecommand{\newblock}{\relax}
\providecommand{\bibinfo}[2]{#2}
\providecommand{\BIBentrySTDinterwordspacing}{\spaceskip=0pt\relax}
\providecommand{\BIBentryALTinterwordstretchfactor}{4}
\providecommand{\BIBentryALTinterwordspacing}{\spaceskip=\fontdimen2\font plus
\BIBentryALTinterwordstretchfactor\fontdimen3\font minus \fontdimen4\font\relax}
\providecommand{\BIBforeignlanguage}[2]{{%
\expandafter\ifx\csname l@#1\endcsname\relax
\typeout{** WARNING: IEEEtran.bst: No hyphenation pattern has been}%
\typeout{** loaded for the language `#1'. Using the pattern for}%
\typeout{** the default language instead.}%
\else
\language=\csname l@#1\endcsname
\fi
#2}}
\providecommand{\BIBdecl}{\relax}
\BIBdecl

\bibitem{xiao2020timme}
Z.~Xiao, W.~Song, H.~Xu, Z.~Ren, and Y.~Sun, ``Timme: Twitter ideology-detection via multi-task multi-relational embedding,'' in \emph{KDD}, 2020, pp. 2258--2268.

\bibitem{yang2022hierarchicaloverlappingbeliefestimation}
\BIBentryALTinterwordspacing
C.~Yang, J.~Li, R.~Wang, S.~Yao, H.~Shao, D.~Liu, S.~Liu, T.~Wang, and T.~F. Abdelzaher, ``Hierarchical overlapping belief estimation by structured matrix factorization,'' 2022. [Online]. Available: \url{https://arxiv.org/abs/2002.05797}
\BIBentrySTDinterwordspacing

\bibitem{li2022unsupervised}
J.~Li, H.~Shao, D.~Sun, R.~Wang, Y.~Yan, J.~Li, S.~Liu, H.~Tong, and T.~Abdelzaher, ``Unsupervised belief representation learning with information-theoretic variational graph auto-encoders,'' in \emph{Proceedings of the 45th International ACM SIGIR Conference on Research and Development in Information Retrieval}, 2022, pp. 1728--1738.

\bibitem{10.3389/fdata.2021.729881}
\BIBentryALTinterwordspacing
D.~Sun, C.~Yang, J.~Li, R.~Wang, S.~Yao, H.~Shao, D.~Liu, S.~Liu, T.~Wang, and T.~F. Abdelzaher, ``Computational modeling of hierarchically polarized groups by structured matrix factorization,'' \emph{Frontiers in Big Data}, vol.~4, 2021. [Online]. Available: \url{https://www.frontiersin.org/journals/big-data/articles/10.3389/fdata.2021.729881}
\BIBentrySTDinterwordspacing

\bibitem{li2024large}
J.~Li, R.~Han, C.~Sun, D.~Sun, R.~Wang, J.~Zeng, Y.~Yan, H.~Tong, and T.~Abdelzaher, ``Large language model-guided disentangled belief representation learning on polarized social graphs,'' in \emph{2024 33rd International Conference on Computer Communications and Networks (ICCCN)}.\hskip 1em plus 0.5em minus 0.4em\relax IEEE, 2024, pp. 1--9.

\bibitem{dong2017weakly}
R.~Dong, Y.~Sun, L.~Wang, Y.~Gu, and Y.~Zhong, ``Weakly-guided user stance prediction via joint modeling of content and social interaction,'' in \emph{Proceedings of the 2017 ACM on Conference on Information and Knowledge Management}, 2017, pp. 1249--1258.

\bibitem{al2017unveiling}
M.~T. Al~Amin, C.~Aggarwal, S.~Yao, T.~Abdelzaher, and L.~Kaplan, ``Unveiling polarization in social networks: A matrix factorization approach,'' in \emph{IEEE INFOCOM 2017-IEEE Conference on Computer Communications}.\hskip 1em plus 0.5em minus 0.4em\relax IEEE, 2017, pp. 1--9.

\bibitem{liu2023unsupervisedimageclassificationideological}
\BIBentryALTinterwordspacing
X.~Liu, J.~Li, D.~Sun, R.~Wang, T.~Abdelzaher, M.~Brown, A.~Barricelli, M.~Kirchner, and A.~Basharat, ``Unsupervised image classification by ideological affiliation from user-content interaction patterns,'' 2023. [Online]. Available: \url{https://arxiv.org/abs/2305.14494}
\BIBentrySTDinterwordspacing

\bibitem{10429919}
X.~Liu, R.~Wang, D.~Sun, J.~Li, C.~Youn, Y.~Lyu, J.~Zhan, D.~Wu, X.~Xu, M.~Liu, X.~Lei, Z.~Xu, Y.~Zhang, Z.~Li, Q.~Yang, and T.~Abdelzaher, ``Influence pathway discovery on social media,'' in \emph{2023 IEEE 9th International Conference on Collaboration and Internet Computing (CIC)}, 2023, pp. 105--109.

\bibitem{settles2009active}
\BIBentryALTinterwordspacing
B.~Settles, ``Active learning literature survey,'' University of Wisconsin--Madison, Computer Sciences Technical Report 1648, 2009. [Online]. Available: \url{http://axon.cs.byu.edu/~martinez/classes/778/Papers/settles.activelearning.pdf}
\BIBentrySTDinterwordspacing

\bibitem{cai2017active}
H.~Cai, V.~W. Zheng, and K.~C.-C. Chang, ``Active learning for graph embedding,'' \emph{arXiv preprint arXiv:1705.05085}, 2017.

\bibitem{chen2019activehneactiveheterogeneousnetwork}
\BIBentryALTinterwordspacing
X.~Chen, G.~Yu, J.~Wang, C.~Domeniconi, Z.~Li, and X.~Zhang, ``Activehne: Active heterogeneous network embedding,'' 2019. [Online]. Available: \url{https://arxiv.org/abs/1905.05659}
\BIBentrySTDinterwordspacing

\bibitem{anrmab}
L.~Gao, H.~Yang, C.~Zhou, J.~Wu, S.~Pan, and Y.~Hu, ``Active discriminative network representation learning,'' in \emph{Proceedings of the 27th International Joint Conference on Artificial Intelligence}, ser. IJCAI'18.\hskip 1em plus 0.5em minus 0.4em\relax AAAI Press, 2018, p. 2142–2148.

\bibitem{zhao2021data}
T.~Zhao, Y.~Liu, L.~Neves, O.~Woodford, M.~Jiang, and N.~Shah, ``Data augmentation for graph neural networks,'' in \emph{Proceedings of the aaai conference on artificial intelligence}, vol.~35, no.~12, 2021, pp. 11\,015--11\,023.

\bibitem{freeman1977set}
L.~C. Freeman, ``A set of measures of centrality based on betweenness,'' \emph{Sociometry}, pp. 35--41, 1977.

\bibitem{pagerankalg}
M.~A. Rodriguez, ``Grammar-based random walkers in semantic networks,'' \emph{Knowledge-Based Systems}, vol.~21, no.~7, pp. 727--739, 2008.

\bibitem{scorenormal}
Y.~Zhang, M.~Lease, and B.~Wallace, ``Active discriminative text representation learning,'' in \emph{Proceedings of the AAAI conference on artificial intelligence}, vol.~31, no.~1, 2017.

\bibitem{zhang2017active}
------, ``Active discriminative text representation learning,'' in \emph{Proceedings of the AAAI conference on artificial intelligence}, vol.~31, no.~1, 2017.

\bibitem{graphcontrastive}
Y.~You, T.~Chen, Y.~Sui, T.~Chen, Z.~Wang, and Y.~Shen, ``Graph contrastive learning with augmentations,'' \emph{Advances in neural information processing systems}, vol.~33, pp. 5812--5823, 2020.

\bibitem{radford2021learning}
A.~Radford, J.~W. Kim, C.~Hallacy, A.~Ramesh, G.~Goh, S.~Agarwal, G.~Sastry, A.~Askell, P.~Mishkin, J.~Clark \emph{et~al.}, ``Learning transferable visual models from natural language supervision,'' in \emph{International conference on machine learning}.\hskip 1em plus 0.5em minus 0.4em\relax PMLR, 2021, pp. 8748--8763.

\bibitem{zhang-etal-2023-pieclass}
\BIBentryALTinterwordspacing
Y.~Zhang, M.~Jiang, Y.~Meng, Y.~Zhang, and J.~Han, ``{PIEC}lass: Weakly-supervised text classification with prompting and noise-robust iterative ensemble training,'' in \emph{Proceedings of the 2023 Conference on Empirical Methods in Natural Language Processing}, H.~Bouamor, J.~Pino, and K.~Bali, Eds.\hskip 1em plus 0.5em minus 0.4em\relax Singapore: Association for Computational Linguistics, Dec. 2023, pp. 12\,655--12\,670. [Online]. Available: \url{https://aclanthology.org/2023.emnlp-main.780}
\BIBentrySTDinterwordspacing

\bibitem{hu2020graph}
S.~Hu, Z.~Xiong, M.~Qu, X.~Yuan, M.-A. C{\^o}t{\'e}, Z.~Liu, and J.~Tang, ``Graph policy network for transferable active learning on graphs,'' \emph{Advances in Neural Information Processing Systems}, vol.~33, pp. 10\,174--10\,185, 2020.

\bibitem{zhang2021rim}
W.~Zhang, Y.~Wang, Z.~You, M.~Cao, P.~Huang, J.~Shan, Z.~Yang, and B.~Cui, ``Rim: Reliable influence-based active learning on graphs,'' \emph{Advances in Neural Information Processing Systems}, vol.~34, pp. 27\,978--27\,990, 2021.

\bibitem{Kipf:2016tc}
\BIBentryALTinterwordspacing
T.~N. Kipf and M.~Welling, ``{Semi-Supervised Classification with Graph Convolutional Networks},'' in \emph{Proceedings of the 5th International Conference on Learning Representations}, ser. ICLR '17, 2017. [Online]. Available: \url{https://openreview.net/forum?id=SJU4ayYgl}
\BIBentrySTDinterwordspacing

\bibitem{velivckovic2017graph}
P.~Veli{\v{c}}kovi{\'c}, G.~Cucurull, A.~Casanova, A.~Romero, P.~Lio, and Y.~Bengio, ``Graph attention networks,'' \emph{arXiv preprint arXiv:1710.10903}, 2017.

\bibitem{li2022sentiment}
Y.~Li and N.~Li, ``Sentiment analysis of weibo comments based on graph neural network,'' \emph{IEEE Access}, vol.~10, pp. 23\,497--23\,510, 2022.

\bibitem{zarandi2024survey}
A.~K. Zarandi and S.~Mirzaei, ``A survey of aspect-based sentiment analysis classification with a focus on graph neural network methods,'' \emph{Multimedia Tools and Applications}, vol.~83, no.~19, pp. 56\,619--56\,695, 2024.

\bibitem{zhang2024doubleh}
C.~Zhang, Z.~Zhou, X.~Peng, and K.~Xu, ``Doubleh: Twitter user stance detection via bipartite graph neural networks,'' in \emph{Proceedings of the International AAAI Conference on Web and Social Media}, vol.~18, 2024, pp. 1766--1778.

\bibitem{kipf2016variational}
T.~N. Kipf and M.~Welling, ``Variational graph auto-encoders,'' \emph{arXiv preprint arXiv:1611.07308}, 2016.

\bibitem{chen2020graph}
F.~Chen, Y.-C. Wang, B.~Wang, and C.-C.~J. Kuo, ``Graph representation learning: a survey,'' \emph{APSIPA Transactions on Signal and Information Processing}, vol.~9, p. e15, 2020.

\bibitem{bruna2017community}
J.~Bruna and X.~Li, ``Community detection with graph neural networks,'' \emph{stat}, vol. 1050, p.~27, 2017.

\bibitem{jiang2021social}
J.~Jiang, X.~Ren, and E.~Ferrara, ``Social media polarization and echo chambers: A case study of covid-19,'' \emph{arXiv preprint arXiv:2103.10979}, 2021.

\bibitem{fortunato2010community}
S.~Fortunato, ``Community detection in graphs,'' \emph{Physics reports}, vol. 486, no. 3-5, pp. 75--174, 2010.

\bibitem{van2008visualizing}
L.~Van~der Maaten and G.~Hinton, ``Visualizing data using t-sne.'' \emph{Journal of machine learning research}, vol.~9, no.~11, 2008.

\bibitem{huang2020gnnvis}
Y.~Huang, J.~Zhang, Y.~Yang, Z.~Gong, and Z.~Hao, ``Gnnvis: Visualize large-scale data by learning a graph neural network representation,'' in \emph{Proceedings of the 29th ACM International Conference on Information \& Knowledge Management}, 2020, pp. 545--554.

\bibitem{ying2018graph}
R.~Ying, R.~He, K.~Chen, P.~Eksombatchai, W.~L. Hamilton, and J.~Leskovec, ``Graph convolutional neural networks for web-scale recommender systems,'' in \emph{Proceedings of the 24th ACM SIGKDD international conference on knowledge discovery \& data mining}, 2018, pp. 974--983.

\bibitem{velickovic2019deep}
P.~Velickovic, W.~Fedus, W.~L. Hamilton, P.~Li{\`o}, Y.~Bengio, and R.~D. Hjelm, ``Deep graph infomax.'' \emph{ICLR (Poster)}, vol.~2, no.~3, p.~4, 2019.

\bibitem{song2022graph}
Z.~Song, X.~Yang, Z.~Xu, and I.~King, ``Graph-based semi-supervised learning: A comprehensive review,'' \emph{IEEE Transactions on Neural Networks and Learning Systems}, vol.~34, no.~11, pp. 8174--8194, 2022.

\bibitem{kipf2016semi}
T.~N. Kipf and M.~Welling, ``Semi-supervised classification with graph convolutional networks,'' \emph{arXiv preprint arXiv:1609.02907}, 2016.

\bibitem{yang2014survey}
X.~Yang, Y.~Guo, Y.~Liu, and H.~Steck, ``A survey of collaborative filtering based social recommender systems,'' \emph{Computer communications}, vol.~41, pp. 1--10, 2014.

\bibitem{pham2011clustering}
M.~C. Pham, Y.~Cao, R.~Klamma, and M.~Jarke, ``A clustering approach for collaborative filtering recommendation using social network analysis.'' \emph{J. Univers. Comput. Sci.}, vol.~17, no.~4, pp. 583--604, 2011.

\bibitem{perozzi2014deepwalk}
B.~Perozzi, R.~Al-Rfou, and S.~Skiena, ``Deepwalk: Online learning of social representations,'' in \emph{Proceedings of the 20th ACM SIGKDD international conference on Knowledge discovery and data mining}, 2014, pp. 701--710.

\bibitem{tang2015line}
J.~Tang, M.~Qu, M.~Wang, M.~Zhang, J.~Yan, and Q.~Mei, ``Line: Large-scale information network embedding,'' in \emph{Proceedings of the 24th international conference on world wide web}, 2015, pp. 1067--1077.

\bibitem{li2018deeper}
Q.~Li, Z.~Han, and X.-M. Wu, ``Deeper insights into graph convolutional networks for semi-supervised learning,'' in \emph{Proceedings of the AAAI conference on artificial intelligence}, vol.~32, no.~1, 2018.

\bibitem{pan2019learning}
S.~Pan, R.~Hu, S.-f. Fung, G.~Long, J.~Jiang, and C.~Zhang, ``Learning graph embedding with adversarial training methods,'' \emph{IEEE transactions on cybernetics}, vol.~50, no.~6, pp. 2475--2487, 2019.

\bibitem{shen-etal-2017-deep}
\BIBentryALTinterwordspacing
Y.~Shen, H.~Yun, Z.~Lipton, Y.~Kronrod, and A.~Anandkumar, ``Deep active learning for named entity recognition,'' in \emph{Proceedings of the 2nd Workshop on Representation Learning for {NLP}}, P.~Blunsom, A.~Bordes, K.~Cho, S.~Cohen, C.~Dyer, E.~Grefenstette, K.~M. Hermann, L.~Rimell, J.~Weston, and S.~Yih, Eds.\hskip 1em plus 0.5em minus 0.4em\relax Vancouver, Canada: Association for Computational Linguistics, Aug. 2017, pp. 252--256. [Online]. Available: \url{https://aclanthology.org/W17-2630}
\BIBentrySTDinterwordspacing

\bibitem{9}
\BIBentryALTinterwordspacing
Y.~Gal, R.~Islam, and Z.~Ghahramani, ``Deep {B}ayesian active learning with image data,'' in \emph{Proceedings of the 34th International Conference on Machine Learning}, ser. Proceedings of Machine Learning Research, D.~Precup and Y.~W. Teh, Eds., vol.~70.\hskip 1em plus 0.5em minus 0.4em\relax PMLR, 06--11 Aug 2017, pp. 1183--1192. [Online]. Available: \url{https://proceedings.mlr.press/v70/gal17a.html}
\BIBentrySTDinterwordspacing

\bibitem{10.1145/2623330.2623760}
\BIBentryALTinterwordspacing
A.~Gadde, A.~Anis, and A.~Ortega, ``Active semi-supervised learning using sampling theory for graph signals,'' in \emph{Proceedings of the 20th ACM SIGKDD International Conference on Knowledge Discovery and Data Mining}, ser. KDD '14.\hskip 1em plus 0.5em minus 0.4em\relax New York, NY, USA: Association for Computing Machinery, 2014, p. 492–501. [Online]. Available: \url{https://doi.org/10.1145/2623330.2623760}
\BIBentrySTDinterwordspacing

\bibitem{KDD-2013-GuALH}
Q.~Gu, C.~C. Aggarwal, J.~Liu, and J.~Han, ``{Selective sampling on graphs for classification},'' in \emph{{Proceedings of the 19th International Conference on Knowledge Discovery and Data Mining}}.\hskip 1em plus 0.5em minus 0.4em\relax {ACM}, 2013, pp. 131--139.

\end{thebibliography}

\end{document}